\documentclass[10pt,twocolumn,letterpaper]{article}

\usepackage[pagenumbers]{cvpr} 

\usepackage[dvipsnames]{xcolor}

\usepackage{appendix}
\usepackage{titling}
    \pretitle{\vspace*{-3pt}\begin{center}\Large \bf}
    \posttitle{\vspace*{10pt}\par\end{center}}
    \preauthor{\large\begin{center}\par}
    \postauthor{\vspace*{6pt}\par\end{center}}
    \predate{}
    \date{}
    \postdate{}

\usepackage{graphicx}
\usepackage{amsmath}
\usepackage{amssymb}
\usepackage{booktabs}
\usepackage{algorithm}
\usepackage[noend]{algpseudocode}
\usepackage{multirow}
\usepackage{enumerate}
\usepackage{comment}
\usepackage{stackengine}
\usepackage{gensymb}
\usepackage{pifont}
\usepackage[accsupp]{axessibility}
\usepackage{tabularx}
\newcommand*\samethanks[1][\value{footnote}]{\footnotemark[#1]}
\usepackage{subcaption}
\usepackage{xcolor} 
\usepackage{accents}
\usepackage{times}
\usepackage{tikz}
\usepackage{mathtools}
\usepackage[super]{nth}

\definecolor{cvprblue}{rgb}{0.21,0.49,0.74}
\usepackage[pagebackref,breaklinks,colorlinks,citecolor=cvprblue]{hyperref}

\usepackage[capitalize]{cleveref}
\crefname{section}{Sec.}{Secs.}
\Crefname{section}{Section}{Sections}
\Crefname{table}{Table}{Tables}
\crefname{table}{Tab.}{Tabs.}

\newlength{\dhatheight}

\newcommand\cmark[1][]{%
  \tikz[scale=0.4,#1]{\fill(0,.35) -- (.25,0) -- (1,.7) -- (.25,.15) -- cycle;}%
}
\newcommand\crossmark[1][]{%
  \tikz[scale=0.4,#1]{
    \fill(0,0)--(0.1,0) .. controls (0.5,0.4) .. (1,0.7)--(0.9,0.7) ..  controls (0.5,0.5) ..(0,0.1) --cycle;
    \fill(1,0.1)--(0.9,0.1) .. controls (0.5,0.3) .. (0,0.7)--(0.1,0.7) .. controls (0.5,0.4) ..(1,0.2) --cycle;
  }%
}

\title{CNC-Net: Self-Supervised Learning for CNC Machining Operations}

\author{Mohsen Yavartanoo$^{1}$\thanks{equal contribution} \qquad Sangmin Hong$^{2}$\samethanks \qquad Reyhaneh Neshatavar$^{1}$\samethanks \qquad Kyoung Mu Lee$^{1,2}$ \\$^{1}$Dept. of ECE \& ASRI, $^{2}$IPAI, Seoul National University, Seoul, Korea\\
{\tt\small \{myavartanoo,mchiash2,reyhanehneshat,kyoungmu\}@snu.ac.kr}}

\begin{document}

\maketitle

\begin{abstract}
CNC manufacturing is a process that employs computer numerical control~(CNC) machines to govern the movements of various industrial tools and machinery, encompassing equipment ranging from grinders and lathes to mills and CNC routers.
However, the reliance on manual CNC programming has become a bottleneck, and the requirement for expert knowledge can result in significant costs.
Therefore, we introduce a pioneering approach named CNC-Net, representing the use of deep neural networks~(DNNs) to simulate CNC machines and grasp intricate operations when supplied with raw materials.
CNC-Net constitutes a self-supervised framework that exclusively takes an input 3D model and subsequently generates the essential operation parameters required by the CNC machine to construct the object.
Our method has the potential to transformative automation in manufacturing by offering a cost-effective alternative to the high costs of manual CNC programming while maintaining exceptional precision in 3D object production.
Our experiments underscore the effectiveness of our CNC-Net in constructing the desired 3D objects through the utilization of CNC operations. 
Notably, it excels in preserving finer local details, exhibiting a marked enhancement in precision compared to the state-of-the-art 3D CAD reconstruction approaches.
\end{abstract}

\section{Introduction}
%
%
%
%
%
Manufacturing processes have undergone remarkable transformations over the past decades, driven by automation and the advancement of computational techniques. 
A domain that has witnessed substantial innovation is Computer Numerical Control~(CNC) machining, a pivotal pillar of modern manufacturing. 
CNC machines have revolutionized manufacturing by producing complex products with better precision, efficiency, and robustness~\cite{altintas2001manufacturing} in diverse industries, from aerospace to medical devices.
Despite their numerous advantages, CNC machines still grapple with certain limitations, particularly in manual programming and adaptability.
Traditional CNC programming requires intricate sets of instructions crafted by Computer-Aided Manufacturing~(CAM) software that guide machine tools, including mills and drills, to produce the intended object. 
However, despite its effectiveness, this process introduces bottlenecks due to its labor-intensive nature and reliance on expert knowledge. 
Furthermore, adapting CNC machines to new tasks typically involves extensive reprogramming, hindering their agility and responsiveness in dynamic manufacturing environments.
Incorporating deep learning techniques into CNC machining offers a transformative solution to address these challenges. 
%
%
In particular, several recent studies use deep neural networks (DNNs) to explore 3D objects using Constructive Solid Geometry~(CSG)~\cite{CSG} operations, employing both a set of simple~\cite{tulsiani2017learning, Csgnet} and more complex~\cite{Cvxnet, 3dias, CAPRI-Net} primitives.
Therefore, the ability of DNNs to learn complex patterns from data makes them an ideal candidate for revolutionizing CNC manufacturing, which can pave the way for automation, adaptive programming, and efficient utilization of CNC machines.
However, the intricate search space for operations on complex objects involving NP-hard problems presents a challenge in labeling optimal solutions as ground truth~(GT).
Consequently, lacking a dataset with such a GT as supervision poses challenges in training a DNN model.

To mitigate these challenges, we propose CNC-Net, a DNN-based approach designed to simulate generic CNC machines in a self-supervised manner.
Our approach can construct target objects without relying on the GT labels~(\ie, a set of sequential operations).
CNC-Net is structured to incrementally learn the production of target 3D shapes, thereby determining the subsequent set of operations by implicitly modeling milling and drilling operations. 
This capability enables CNC-Net to generate the necessary machining steps effectively.
At each operational step, the tools are represented as cylindric primitives, and the CNC-Net determines the radius of the tool and identifies the path coordinates for the subsequent milling or drilling action.
To enhance the carving capabilities of a CNC machine, we introduced a feature that enables the machine to rotate the workpiece along the $X$ and $Y$ axes.
This functionality is commonly found in advanced CNC setups. 
In this scenario, guided solely by the target shape and lacking prior information or labeled operations, CNC-Net models 3D shapes at each step, involving subtracting operations, represented as the union of cylindric primitives, from the outcomes of preceding steps.
This enables CNC-Net to learn the essential operations to reconstruct the target shapes accurately.

In our experiments, we provide the competitive performance of CNC-Net in reproducing 3D shapes compared to state-of-the-art~(SOTA) CAD reconstruction methods.
To validate the effectiveness of our approach, we conduct experiments on both industrial objects from the ABC dataset~\cite{Abc} and more intricate objects obtained from the ShapeNet dataset~\cite{Shapenet}.
Our main contributions are threefold:
\begin{itemize}
\item We introduce CNC-Net, a pioneer self-supervised and DNN-based approach for simulating CNC machines.
\item CNC-Net learns to automatically find the sequential operations required for sculpting a 3D shape and exhibits capability akin to expert human labor without the need for labels or any prior information.
\item The experiments demonstrate that our self-supervised CNC-Net method can precisely reproduce target objects and outperform SOTA methods in terms of 3D reconstruction performance based on volume-based metrics.
\end{itemize}

\begin{figure*}[t]
\centering
\begin{subfigure}[b]{0.13\textwidth}
    \includegraphics[width=\linewidth, page=1]{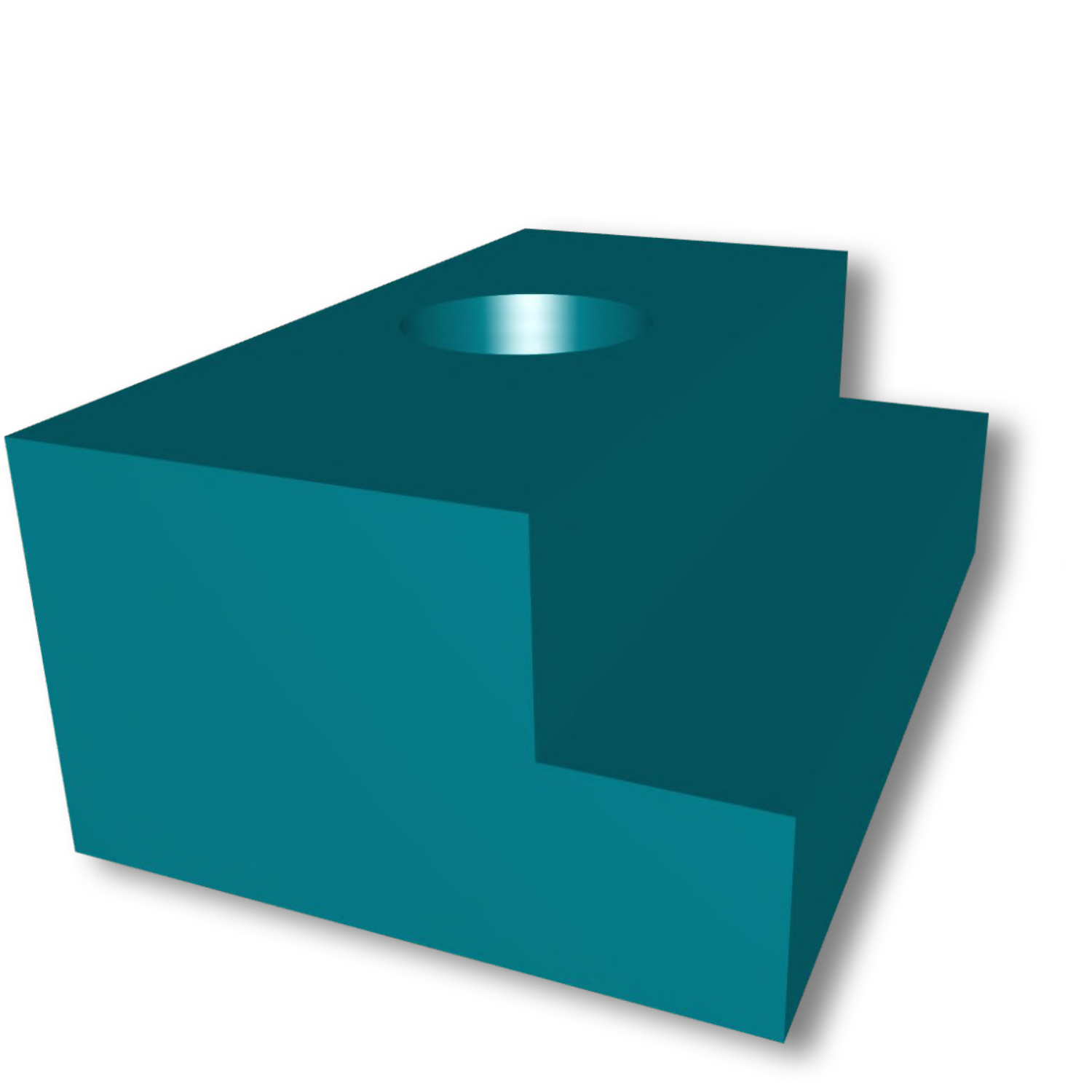}
    \caption{
        \textbf{Target Shape.}
    }
    \label{fig:CNC_target}
\end{subfigure}
\begin{subfigure}[b]{0.13\textwidth}
        \includegraphics[width=\linewidth, page=2]{figures/Operations.pdf}
    \caption{
        \textbf{Input Material.}
    }
    \label{fig:CNC_material}
\end{subfigure}
\begin{subfigure}[b]{0.13\textwidth}
        \includegraphics[width=\linewidth, page=3]{figures/Operations.pdf}
    \caption{
        \textbf{Tool.}
    }
    \label{fig:CNC_tool}
\end{subfigure}
\begin{subfigure}[b]{0.13\textwidth}
        \includegraphics[width=\linewidth, page=4]{figures/Operations.pdf}
    \caption{
        \textbf{Path.}
    }
    \label{fig:CNC_path}
\end{subfigure}
\begin{subfigure}[b]{0.13\textwidth}
        \includegraphics[width=\linewidth, page=5]{figures/Operations.pdf}
    \caption{
        \textbf{Milling.}
    }
    \label{fig:CNC_mill}
\end{subfigure}
\begin{subfigure}[b]{0.13\textwidth}
        \includegraphics[width=\linewidth, page=6]{figures/Operations.pdf}
    \caption{
        \textbf{Drilling.}
    }
    \label{fig:CNC_drill}
\end{subfigure}
\begin{subfigure}[b]{0.13\textwidth}
        \includegraphics[width=\linewidth, page=7]{figures/Operations.pdf}
    \caption{
        \textbf{Rotation.}
    }
    \label{fig:CNC_rot}
\end{subfigure}
    \caption{\textbf{Overview of a generic CNC machine features.}}
    \label{fig:CNC}
    \vspace{-4mm}
\end{figure*}

\section{Related Works}
This section covers previous studies related to our method, divided into two categories: reverse engineering of 3D shapes and machine learning for CNC machines.
\paragraph{Reverse engineering 3d shapes.}
Reverse engineering a 3D shape refers to understanding the features and structure of the original object and learning how it is constructed. 
With the development of deep learning, 
several approaches have been proposed to investigate how a 3D shape is assembled.
%
In recent years, there has been an exploration of the use of simple geometric primitives to approximate a 3D shape with a pre-defined set of cubes~\cite{Tulsiani_2017_CVPR, Zou_2017_ICCV,  niu2018im2struct}, ellipsoids~\cite{genova2019learning}.
More recent studies improve the representation ability and surface reconstruction by introducing more flexible and deformable primitives~\cite{Cvxnet, 3dias, Neural_parts, huang2023learning}.
These works represent a shape as a union of primitives using constructive solid geometry~(CSG)~\cite{CSG}, which relies on Boolean operations applied to the primitives~\cite{foley1996computer}.
%
%
On the other hand, there exist various methods~\cite{Csgnet, du2018, Brick-by-brick} that assemble primitives using a sequence of modeling operations through reinforcement learning~(RL).
These methods aim to match a target shape in a reverse engineering manner.
Furthermore, recent supervised primitive networks~\cite{SPFN, ParseNet} have been designed to detect and fit primitives within point clouds, which initially identify primitive types.
Subsequently, they estimate their parameters or integrate spline patches, incorporating differentiable metric-learning segmentation.
Additionally, CSGNet~\cite{Csgnet} is a neural network approach to form a CSG program from a given shape, and InverseCSG~\cite{du2018} solves it as a program synthesis problem.
Later methods~\cite{UCSG, CSG-Stump, ExtrudeNet, CAPRI-Net, SECAD-Net} learn to compact 3D computer-aided design~(CAD) models via CSG operations, including intersection, union, and subtraction, without relying on any ground-truth primitive assemblies.
However, primitive and implicit-based methods are often designed for static shape reconstruction rather than dynamic processes such as material removal in machining.
Accordingly, none of the above-mentioned methods is applicable for modeling operations in CNC machining. 
%
%
In contrast, our self-supervised method can dynamically reproduce target shapes by learning sequential CNC machining operations.
\paragraph{Machine learning for CNC machines.}
%
%
Computer-aided process planning (CAPP) by utilizing machine learning~(ML) approaches plays a crucial role in streamlining and automating various stages of the manufacturing process~\cite{komura2023computer, leo2015feature}.
It encompasses critical machining processes such as toolpath optimization, feature selection, 
tool selection, operation selection, etc.
%
ML-based methods~\cite{DITTRICH201949, HSIEH20133, zavalnyi2019optimization, natarajan2021, von2022machine, li2020trajectory} within CAPP have focused on improving toolpath generation and optimization, contributing to manufacturing processes. 
In~\cite{HSIEH20133}, particle swarm optimization~(PSO) is utilized to optimize randomly initialized toolpaths to minimize the machining time and resource consumption.
Furthermore, a recent study~\cite{DITTRICH201949} utilizes support vector machine~(SVM) for error prediction in toolpath generation, ensuring that the toolpaths generated meet the quality and precision requirements.
%
%
Such optimization-based methods enhance the machining performance and reduce production costs, but they lack the capability to learn the CNC operations themselves.
Moreover, \cite{BALIC20021171, kukreja2022optimum} has been proposed to select the best route planning strategy from conventional approaches, helping to generate a path aligned with specific manufacturing objectives.
However, these methods require data preparation involving 3D CAD models, each with corresponding labels representing the generated toolpaths, which is time-consuming.
Moreover, such models are limited to learning only from these specific labels and restricting their generalization to adapt and generate toolpaths across different datasets.
Consequently, direct research on automatic search and learning to generate CNC machining operations step-by-step has been relatively scarce. 
Therefore, we introduce a novel self-supervised framework, CNC-Net, to simulate generic CNC machines by generating toolpaths and learning the operations sequentially solely from 3D CAD models without requiring any labeled data or prior information.
Our proposed method can be applied to diverse datasets containing 3D CAD models and shows the ability to learn the operations, even from a single sample, using zero-shot learning.
\\

\section{Method}
In this section, we first provide an overview of a generic CNC machine by explaining its basic principles. 
Later, we introduce our novel framework, CNC-Net, which takes an input 3D model and attempts to generate the sequence of operations required for the CNC machine to reproduce the designated shape in a fully self-supervised manner.
\subsection{CNC machine}
Computer Numerical Control~(CNC) is a computer-based system used to control various machining tools such as drills and mills. 
It operates using pre-programmed instructions, such as the G-code or M-code, to shape materials in the desired shape $\mathcal{S}$ as~\cref{fig:CNC_target}.
These instructions include a sequence of operations that can be created by individuals, computer-aided design~(CAD) systems, or computer-aided manufacturing~(CAM) software.
In this study, we only consider CNC machines that use milling and drilling processes applied from the top of the workpiece and include rotation capability as a typical machining operation.
%
\subsubsection{Input material}
As shown in~\cref{fig:CNC_material}, we implicitly model the material initially provided, denoted as $\mathcal{S}_0$, as the bounding box $\mathcal{B}(\mathcal{S})$ that encompasses the target object $\mathcal{S}$ as follows:
\begin{equation}\label{eq:cube}
\begin{split}
\mathcal{S}_0=\mathcal{B}(\mathcal{S})=\textbf{max}(|\frac{x}{l}|,|\frac{y}{w}|,|\frac{z}{h}|)-1,
\end{split}
\end{equation}
where $l$, $w$, and $h$ represent the length, width, and height of the target object $\mathcal{S}$, respectively.
\subsubsection{Tool}
As shown in~\cref{fig:CNC_tool}, for simplicity and without loss of generality, we model milling and drilling tools $\mathcal{T}$ using the implicit form of cylindrical primitives represented as follows:
\begin{equation}\label{eq:milling_tool}
\mathcal{T}(c,r)=\textbf{max}\{(x-c_x)^2+(y-c_y)^2-r^2,c_z-z\},
\end{equation}
where $r$ and $c=(c_x,c_y,c_z)$ denote the radius of the tool and the center of the base of the cylinder, respectively.
\subsubsection{Path}
%
\begin{figure*}[t]
\centering
    \includegraphics[width=\linewidth]{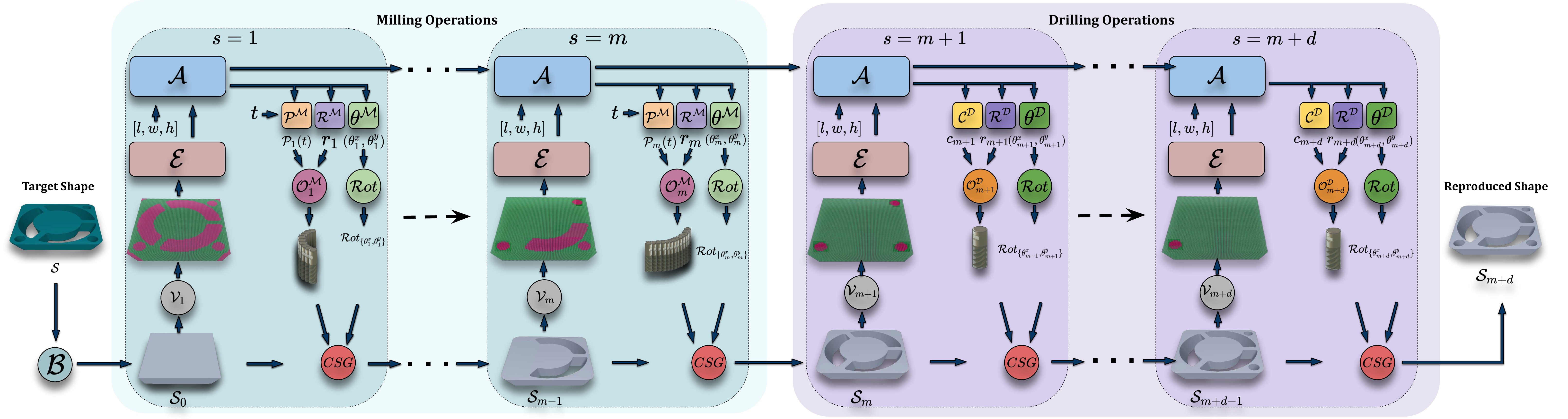}
    \caption{
        \textbf{Overview of our proposed framework.} Subsequent of milling and drilling operations to reproduce a target 3D CAD model.
    }
    \label{fig:framework}
    \vspace{-4mm}
\end{figure*}
The milling tool can move along a specified path to carve or cut through the material.
As shown in~\cref{fig:CNC_path}, we represent the path using a parametric function $\mathcal{P}$ as follows:
\begin{equation}\label{eq:path}
\begin{split}
\mathcal{P}(t)=(c_x(t),c_y(t),c_z),
\end{split}
\end{equation}
where $c_x(t)$ and $c_y(t)$ represent the parametric components for each axis X and Y, with the parameter $t$, respectively. 
It should be noted that the component $c_z$ is constant, indicating the depth of penetration of the tool into the workpiece.
\subsubsection{Milling operation}
Milling is the process of removing material from a workpiece using a rotating cutting tool, often with multiple edges layer by layer, as shown in~\cref{fig:CNC_mill}. 
It is commonly used to create complex contours, pockets, and slots.
Consequently, milling operation $\mathcal{O}^{\mathcal{M}}_s$ at step $s$ can be defined as the process of a tool $\mathcal{T}_s$ traversing a path $\mathcal{P}_s(t)$, expressed as follows:
\begin{equation}\label{eq:milling_op}
\begin{split}
\mathcal{O}^{\mathcal{M}}_s=\textbf{min}\{\mathcal{T}_s(\mathcal{P}_s(t),r_s)\}_{t=0}^{1},
\end{split}
\end{equation}
where $r_s$ is the radius of the tool at step $s$ and the $\textbf{min}$ operator encompasses the union of the primitives $\mathcal{T}_s$ across path parameterized by time steps $t$ ranging from $0$ to $1$.

\subsubsection{Drilling operation}
Drilling is a machining process that uses a rotating drill bit to create holes with specific sizes and depths in a workpiece. 
This operation is essential to accommodate fasteners such as bolts and screws in various applications.
Unlike the milling operation, which traverses along a specific path, the drilling tool $\mathcal{T}_s$ at step $s$ penetrates a designated location $c_s=(c_{s,x}, c_{s,y}, c_{s,z})$ as shown in~\cref{fig:CNC_drill}.
As a result, the drilling operation $\mathcal{O}^{\mathcal{D}}_s$ at step $s$ can be defined as follows:
\begin{equation}\label{eq:drilling_op}
\begin{split}
\mathcal{O}^{\mathcal{D}}_s=\mathcal{T}_s(c_s,r_s),
\end{split}
\end{equation}
where $r_s$ represents the radius of the drilling tool at step $s$. 
\subsubsection{Rotation operation}
CNC machines are equipped with the ability to rotate the workpiece at different angles, enhancing their versatility, precision, and effectiveness in addressing complex geometries and intricate machining tasks while eliminating the need for manual repositioning.
As shown in~\cref{fig:CNC_rot}, we assume that the machine can rotate the shape $\mathcal{S}_s$ at step $s$ by the rotation transformation $\mathcal{R}ot_{\{\theta^x_s,\theta^y_s\}}$ as follows:
\begin{equation}\label{eq:rotation}
\begin{split}
\mathcal{S}^{\mathcal{R}ot}_s &= \mathcal{R}ot_{\{\theta^x_s,\theta^y_s\}}(\mathcal{S}_s),\\
\end{split}
\end{equation}
where $\theta^x_s$ and $\theta^y_s$ denote the counterclockwise rotation angles about the $X$-axis and $Y$-axis at step $s$, respectively.

\subsection{CNC-Net}
In this section, we introduce CNC-Net, an innovative self-supervised framework designed to automate the operations of generic CNC machines. 
CNC-Net achieves this by autonomously learning and refining essential parameters for sequential operations, including milling, drilling, and rotation, to sculpt raw materials into intricate objects precisely.
\cref{fig:framework} provides a comprehensive visual representation of our proposed self-supervised framework, CNC-Net.
We start with voxelizing the input material $\mathcal{S}_0$ as the bounding box $\mathcal{B}(\mathcal{S})$ of the target shape $\mathcal{S}$ as follows:
\begin{equation}\label{eq:vox}
  \mathcal{V}_0(v)=\begin{cases}
    +1, & \text{if $v\in (\mathcal{S}_0-\mathcal{S})$}\\
    -1, & \text{if $v\in \mathcal{S}$}
  \end{cases}.
\end{equation}
%
Subsequently, we utilize an encoder $\mathcal{E}$ to extract both local and global features from $\mathcal{V}_0$, where the encoded features are concatenated with the bounding box size $[l,w,h]$ and fed into a long short-term memory~(LSTM) network denoted as $\mathcal{A}$. 
This LSTM network generates hidden features for the subsequent operational step and produces the necessary output features to predict the parameters of the operations.
Moreover, we utilize three distinct decoders $\theta^{\mathcal{M}}$, $\mathcal{R}^{\mathcal{M}}$, and $\mathcal{P}^{\mathcal{M}}$ to generate the rotation parameters $(\theta^x
_1,\theta^y_1)$, tool radius $r_1$, and path $\mathcal{P}_1(t)$ parameterized by $t$ for the first step $s=1$, respectively.
Then, we construct the milling operation $\mathcal{O}_1^{\mathcal{M}}$ from the generated $r_1$ and $\mathcal{P}_1(t)$ and feed it along with $\mathcal{S}_0$ and the rotation parameters $(\theta^x
_1,\theta^y_1)$ into the Constructive Solid Geometry (CSG) operation, executing $\mathcal{O}^{\mathcal{M}}_1$ and the rotation $\mathcal{R}ot_{\{\theta^x
_1,\theta^y_1\}}$ resulting in the generation of $\mathcal{S}_1$.
We continue this process with milling operations $\mathcal{O}^{\mathcal{M}}_s$ iteratively for $m$ steps until no further improvement is achieved and generate the shape $\mathcal{S}_m$ at step $s=m$, wherein we update the voxel representation as follows:
\begin{equation}\label{eq:vox_s}
  \mathcal{V}_{s-1}(v)=\begin{cases}
    +1, & \text{if $v\in  (\mathcal{S}_{s-1}-\mathcal{S})$ }\\
    -1, & \text{if $v\in \mathcal{S}\cup(\mathcal{S}_0-\mathcal{S}_{s-1})$}
  \end{cases}.
\end{equation}
%
%
%

When the milling operations are completed, we proceed to perform the drilling operations similarly, where the generated shape $\mathcal{S}_m$ is considered as the initial material for the drilling operations. 
Here, we input the updated voxel representation $\mathcal{V}_m$ into the same encoder $\mathcal{E}$ and feed the extracted features along with the bounding box size $[l,w,h]$ to the LSTM network $\mathcal{A}$ to extract the characteristics, but different decoders $\theta^{\mathcal{D}}$, $\mathcal{R}^{\mathcal{D}}$, and $\mathcal{C}^{\mathcal{D}}$ to produce the rotation parameters $(\theta^x_{m+1},\theta^y_{m+1})$, the drill radius $r_{m+1}$, and the drill tip coordinates $c_{m+1}$, respectively. 
Similar to the milling operation, $\mathcal{S}_{m+1}$ in the subsequent step $s=m+1$ can be generated by feeding $\mathcal{S}_{m}$ along with the constructed drilling operation $\mathcal{O}^{\mathcal{D}}_{m+1}$ and the generated rotation parameters $(\theta^x
_{m+1},\theta^y_{m+1})$ into the CSG operation.
We iterate the drilling operations over $d$ steps until a notable similarity is achieved between $\mathcal{S}_{m+d}$ and the target shape $\mathcal{S}$.

\subsubsection{CSG}
%
\begin{figure}[t]
\centering
\begin{subfigure}[b]{0.115\textwidth}
    \includegraphics[width=\linewidth, page=1]{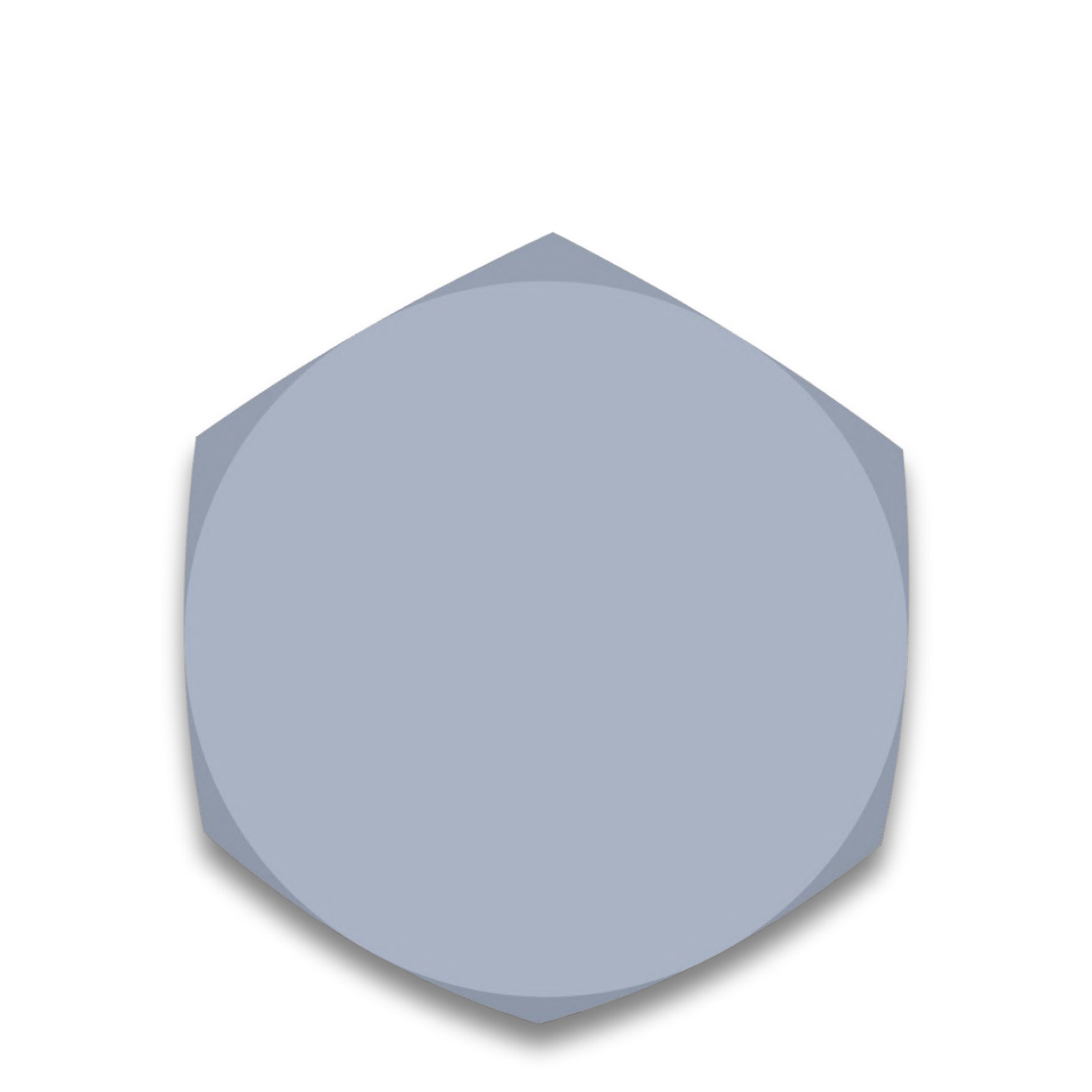}
    \caption{
        \scalebox{.9}{\textbf{\tiny $\mathcal{S}_{s-1}$}}
    }
    \label{fig:CSG_a}
\end{subfigure}
\begin{subfigure}[b]{0.115\textwidth}
        \includegraphics[width=\linewidth, page=2]{figures/CSG_1.pdf}
    \caption{
        \scalebox{.9}{\textbf{\tiny $\mathcal{S}^{\mathcal{R}}_{s-1}$}}
    }
    \label{fig:CSG_b}
\end{subfigure}
\begin{subfigure}[b]{0.115\textwidth}
        \includegraphics[width=\linewidth, page=3]{figures/CSG_1.pdf}
    \caption{
        \scalebox{.9}{\textbf{\tiny $\textbf{max}(\mathcal{S}^{\mathcal{R}}_{s-1},-\mathcal{O}_s)$}}
    }
    \label{fig:CSG_c}
\end{subfigure}
\begin{subfigure}[b]{0.115\textwidth}
        \includegraphics[width=\linewidth, page=4]{figures/CSG_1.pdf}
    \caption{
        \scalebox{.9}{\textbf{\tiny $\mathcal{S}_s$}}
    }
    \label{fig:CSG_d}
\end{subfigure}
    \caption{\textbf{CSG operation.}}
    \label{fig:CSG}
    \vspace{-4mm}
\end{figure}
%
\cref{fig:CSG} shows the outline of the CSG module.
To construct the shape $\mathcal{S}_s$ for each step $s$, the initial procedure involves applying the rotation transformation $\mathcal{R}ot_{\{\theta^x_s,\theta^y_s\}}$ to $\mathcal{S}_{s-1}$ as in ~\cref{eq:rotation}, where the resulting rotated shape is denoted as $\mathcal{S}_{s-1}^{\mathcal{R}ot}$.
%
%
Accordingly, the shape $\mathcal{S}_{s}$ can be constructed through the subtraction of a milling operation $\mathcal{O}^{\mathcal{M}}_s$ or a drilling operation $\mathcal{O}^{\mathcal{D}}_s$ from the rotated shape $\mathcal{S}_{s-1}^{\mathcal{R}ot}$ followed by the inverse rotation $\mathcal{R}ot^{-1}_{\{\theta^x_s,\theta^y_s\}}$ to ensure that its orientation matches the initial orientation as follows:
\begin{equation}\label{eq:csg}
\begin{split}
\mathcal{S}_{s} &= \mathcal{R}ot^{-1}_{\{\theta^x_s,\theta^y_s\}}(\textbf{max}(\mathcal{S}_{s-1}^{\mathcal{R}ot},-\mathcal{O}^{\mathcal{M}}_s))
\\
\mathcal{S}_{s} &= \mathcal{R}ot^{-1}_{\{\theta^x_s,\theta^y_s\}}(\textbf{max}(\mathcal{S}_{s-1}^{\mathcal{R}ot},-\mathcal{O}^{\mathcal{D}}_s))
\end{split} \text{or,}
\end{equation}
where the $\textbf{max}$ operator encompasses the subtraction.

\subsection{Loss functions}\label{sec:losses}
We set up various loss functions for self-supervised training of our model.
For simplicity, we represent $||.||$ as the $L^2$ norm, $v_{xyz}\in\mathbb{R}^{3}$ as the 3D coordinates of the corresponding voxel $v$, $v_{xyz}^{\mathcal{R}ot}\in\mathbb{R}^{3}$ and $v_{xy}^{\mathcal{R}ot}\in\mathbb{R}^{2}$ as the 3D and 2D coordinates of the rotated $v_{xyz}$ by rotation operation $\mathcal{R}ot_{\{\theta^x_s,\theta^y_s\}}$, and $\sigma(x)=\tanh({wx})$ as a smooth sign function, with $w$ serving as a large scaling factor.
First, we design the milling loss $\mathcal{L}^{\mathcal{M}}$ to ensure that the implicit shape $\mathcal{S}_{m}$ obtained after sequential milling precisely approximates the target shape $\mathcal{S}$.
Consequently, $\mathcal{S}_{m}$ is expected to produce negative and positive values for voxels $v\in\mathcal{V}_0$ inside and outside the target shape $\mathcal{S}$ as follows:
\begin{equation}\label{eq:loss_mill}
\begin{split}
\mathcal{L}^{\mathcal{M}}
=
\frac{1}{|\mathcal{V}_0|}\sum_{v\in\mathcal{V}_0}||\sigma(\mathcal{S}_{m}(v_{xyz}))-\mathcal{V}_0(v)||^2,
\end{split}
\end{equation}
where $|\mathcal{V}_0|$ denotes the total number of voxels in $\mathcal{V}_0$.

We further define the drilling loss $\mathcal{L}^{\mathcal{D}}$ to facilitate the drilling operations to remove the remaining regions after the sequential milling operations as follows:
\begin{equation}\label{eq:loss_drill}
\begin{split}
\mathcal{L}^{\mathcal{D}}
=
\frac{1}{d}\smashoperator[l]{\sum_{s=m+1}^{m+d}}\frac{1}{|\mathcal{V}^{+}_{s-1}|}\smashoperator[r]{\sum_{v\in\mathcal{V}^{+}_{s-1}}}||\sigma(\mathcal{O}^{\mathcal{D}}_{s}(v_{xyz}^{\mathcal{R}ot}))+1||^2,
\end{split}
\end{equation}
where $\mathcal{V}^{+}_{s-1}$ is the subset of $\mathcal{V}_{s-1}$ whoes values are positive~($+1$) and $|\mathcal{V}^{+}_{s-1}|$ denotes its number of voxesls. 

To ease training and improve performance, we also introduce a shape loss $\mathcal{L}^{\mathcal{S}}$ and a center loss $\mathcal{L}^{\mathcal{C}}$ as complementary losses for both operations. 
$\mathcal{L}^{\mathcal{S}}$ ensures that the regions inside the target shape $\mathcal{S}$ are not removed by any milling or drilling operations, and therefore $\mathcal{S}$ is preserved after sequential operations.
Therefore, we need to ensure that the milling or drilling operation $\mathcal{O}_s$ at each step $s$ produces positive values for the voxels inside $\mathcal{S}$ as follows:
\begin{equation}\label{eq:loss_shape}
\begin{split}
\mathcal{L}^{\mathcal{S}}
=
\frac{1}{|\mathcal{V}^{-}_0|}\sum_{s=1}^{m+n}\sum_{v\in\mathcal{V}^{-}_0}||\sigma(\mathcal{O}_{s}(v_{xyz}^{\mathcal{R}ot}))-1||^2,
\end{split}
\end{equation}
where $\mathcal{V}^{-}_0$ is the subset of $\mathcal{V}_0$ whose values are negative~($-1$) and $|\mathcal{V}^{-}_0|$ denotes its number of voxels.
Furthermore, $\mathcal{L}^{\mathcal{C}}$ is applied from the top of the workpiece to ensure that the decoders $\mathcal{P}^{\mathcal{M}}$ and $\mathcal{C}^{\mathcal{D}}$ generate tooltip coordinates around the remaining regions that are yet to be removed.
This loss simplifies the training process by narrowing down the search space for the decoders $\mathcal{P}^{\mathcal{M}}$ and $\mathcal{C}^{\mathcal{D}}$, facilitating the generation of tip coordinates.
Consequently, we minimize the Chamfer distance between the 2D coordinates $C_{s}^{xy}$ of the tools tips at each step $s$, \eg, $C_{s}^{xy}=\{(c_{s,x}(t),c_{s,y}(t))\}_{t=0}^{1}$ for the milling path and $C_{s}^{xy}=(c_{s,x},c_{s,y})$ for the drilling tip and the 2D coordinates $v_{xy}^{\mathcal{R}ot}$ of the positive-valued voxels $\mathcal{V}_{s-1}^{+}$ as follows:
\begin{equation}\label{eq:loss_center}
\begin{split}
\mathcal{L}^{\mathcal{C}}
=\frac{1}{m+d}\smashoperator[lr]{\sum_{s=1}^{m+d}}&(
\frac{1}{|C_{s}^{xy}|}\smashoperator[lr]{\sum_{\substack{c_{s}^{xy}\\\in C_{s}^{xy}}}} \smashoperator[r]{\min_{\substack{v\in V_{s-1}^{+}}}}||c^{xy}_s-v^{\mathcal{R}ot}_{xy}||^2
\\+
&\frac{1}{|\mathcal{V}^{+}_{s-1}|}\smashoperator[l]{\sum_{\substack{v\in V_{s-1}^{+}}}}\smashoperator[lr]{\min_{\substack{c_{s}^{xy}\\\in C_{s}^{xy}}}}||v^{\mathcal{R}ot}_{xy}-c^{xy}_s||^2)
\end{split},
\end{equation}
where $|C_{s}^{xy}|$ denotes the number of points inside $C_{s}^{xy}$.
\begin{figure*}[t]
     \centering
     \begin{subfigure}[b]{0.12\textwidth}
        \centering
        \includegraphics[trim={0 0.750cm 0 1.5cm},clip,width=\textwidth, page=1]{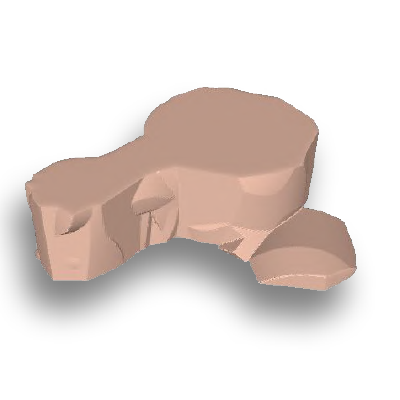}
     \end{subfigure}
     \hfill
     \begin{subfigure}[b]{0.12\textwidth}
        \centering
        \includegraphics[trim={0 0.750cm 0 1.5cm},clip,width=\textwidth, page=2]{figures/ABC_22_1.pdf}
     \end{subfigure}
     \hfill
     \begin{subfigure}[b]{0.12\textwidth}
        \centering
        \includegraphics[trim={0 0.750cm 0 1.5cm},clip,width=\textwidth, page=3]{figures/ABC_22_1.pdf}
     \end{subfigure}
     \hfill
     \begin{subfigure}[b]{0.12\textwidth}
        \centering
        \includegraphics[trim={0 0.750cm 0 1.5cm},clip,width=\textwidth, page=4]{figures/ABC_22_1.pdf}
     \end{subfigure}
     \hfill
     \begin{subfigure}[b]{0.12\textwidth}
        \centering
        \includegraphics[trim={0 0.750cm 0 1.5cm},clip,width=\textwidth, page=6]{figures/ABC_22_1.pdf}
     \end{subfigure}
     \hfill
     \begin{subfigure}[b]{0.12\textwidth}
        \centering
        \includegraphics[trim={0 0.750cm 0 1.5cm},clip,width=\textwidth, page=7]{figures/ABC_22_1.pdf}
     \end{subfigure}
     
     \begin{subfigure}[b]{0.12\textwidth}
        \centering
        \includegraphics[trim={0 0 0 1.5cm},clip, width=\textwidth, page=1]{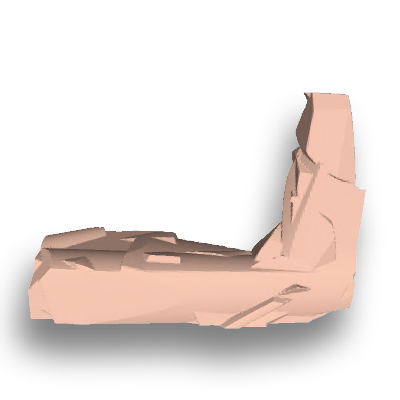}
     \end{subfigure}
     \hfill
     \begin{subfigure}[b]{0.12\textwidth}
        \centering
        \includegraphics[trim={0 0 0 1.5cm},clip,width=\textwidth, page=2]{figures/ABC_377_1.pdf}
     \end{subfigure}
     \hfill
     \begin{subfigure}[b]{0.12\textwidth}
        \centering
        \includegraphics[trim={0 0 0 1.5cm},clip,width=\textwidth, page=3]{figures/ABC_377_1.pdf}
     \end{subfigure}
     \hfill
     \begin{subfigure}[b]{0.12\textwidth}
        \centering
        \includegraphics[trim={0 0 0 1.5cm},clip,width=\textwidth, page=4]{figures/ABC_377_1.pdf}
     \end{subfigure}
     \hfill
     \begin{subfigure}[b]{0.12\textwidth}
        \centering
        \includegraphics[trim={0 0 0 1.5cm},clip,width=\textwidth, page=5]{figures/ABC_377_1.pdf}
     \end{subfigure}
     \hfill
     \begin{subfigure}[b]{0.12\textwidth}
        \centering
        \includegraphics[trim={0 0 0 1.5cm},clip,width=\textwidth, page=7]{figures/ABC_377_1.pdf}
     \end{subfigure}
     
     \begin{subfigure}[b]{0.12\textwidth}
        \centering
        \includegraphics[trim={0 0.75cm 0 1.7cm},clip,width=\textwidth, page=1]{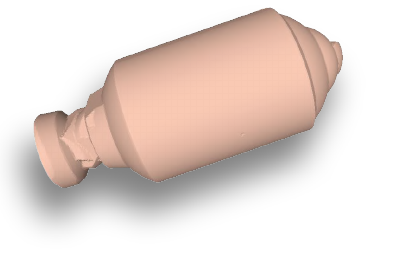}
     \end{subfigure}
     \hfill
     \begin{subfigure}[b]{0.12\textwidth}
        \centering
        \includegraphics[trim={0 0.75cm 0 1.7cm},clip,width=\textwidth, page=2]{figures/ABC_190_1.pdf}
     \end{subfigure}
     \hfill
     \begin{subfigure}[b]{0.12\textwidth}
        \centering
        \includegraphics[trim={0 0.75cm 0 1.7cm},clip,width=\textwidth, page=3]{figures/ABC_190_1.pdf}
     \end{subfigure}
     \hfill
     \begin{subfigure}[b]{0.12\textwidth}
        \centering
        \includegraphics[trim={0 0.75cm 0 1.7cm},clip,width=\textwidth, page=4]{figures/ABC_190_1.pdf}
     \end{subfigure}
     \hfill
     \begin{subfigure}[b]{0.12\textwidth}
        \centering
        \includegraphics[trim={0 0.75cm 0 1.7cm},clip,width=\textwidth, page=5]{figures/ABC_190_1.pdf}
     \end{subfigure}
     \hfill
     \begin{subfigure}[b]{0.12\textwidth}
        \centering
        \includegraphics[trim={0 0.75cm 0 1.7cm},clip,width=\textwidth, page=6]{figures/ABC_190_1.pdf}
     \end{subfigure}
     
     \begin{subfigure}[b]{0.12\textwidth}
        \centering
        \includegraphics[trim={0 0 0 2cm},clip,width=\textwidth, page=1]{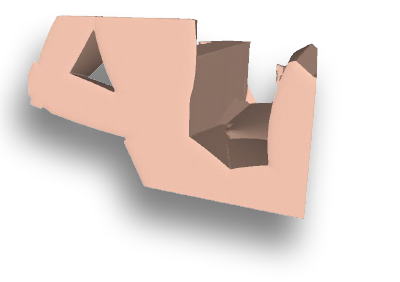}
        \caption*{CSG-Stump~\cite{CSG-Stump}}
     \end{subfigure}
     \hfill
     \begin{subfigure}[b]{0.12\textwidth}
        \centering
        \includegraphics[trim={0 0 0 2cm},clip,width=\textwidth, page=2]{figures/ABC_958_1.pdf}
        \caption*{ExtrudeNet~\cite{ExtrudeNet}}
     \end{subfigure}
     \hfill
     \begin{subfigure}[b]{0.12\textwidth}
        \centering
        \includegraphics[trim={0 0 0 2cm},clip,width=\textwidth, page=3]{figures/ABC_958_1.pdf}
        \caption*{CAPRI-Net~\cite{CAPRI-Net}}
     \end{subfigure}
    \hfill
     \begin{subfigure}[b]{0.12\textwidth}
        \centering
        \includegraphics[trim={0 0 0 2cm},clip,width=\textwidth, page=4]{figures/ABC_958_1.pdf}
        \caption*{SECAD-Net~\cite{SECAD-Net}}
     \end{subfigure}
     \hfill
     \begin{subfigure}[b]{0.12\textwidth}
        \centering
        \includegraphics[trim={0 0 0 2cm},clip,width=\textwidth, page=5]{figures/ABC_958_1.pdf}
        \caption*{CNC-Net~(Ours)}
     \end{subfigure}
     \hfill
     \begin{subfigure}[b]{0.12\textwidth}
        \centering
        \includegraphics[trim={0 0 0 2cm},clip,width=\textwidth, page=6]{figures/ABC_958_1.pdf}
        \caption*{Target~(GT)}
     \end{subfigure}
        \caption{
        \textbf{Qualitative results on ABC~\cite{Abc} dataset.} 
         All reproduced shapes are visualized using marching cubes~(MC) with $256$ resolution.}
        \label{fig:qual_abc}
        \vspace{-4mm}
\end{figure*}

The total loss $\mathcal{L}^{\text{T}}$ is a summation of the defined loss functions $\mathcal{L}^{\mathcal{M}}$, $\mathcal{L}^{\mathcal{D}}$, $\mathcal{L}^{\mathcal{S}}$, and $\mathcal{L}^{\mathcal{C}}$, expressed as follows:
\begin{equation}\label{eq:loss_opt}
\begin{split}
\mathcal{L}^{\text{T}}=
\mathcal{L}^{\mathcal{M}}+
\mathcal{L}^{\mathcal{D}}+
\mathcal{L}^{\mathcal{S}}+
\mathcal{L}^{\mathcal{C}}.
\end{split}
\end{equation}

\subsection{Implementation details}
We utilize the same encoder network of CapriNet~\cite{CAPRI-Net} as $\mathcal{E}$ and a two-layer LSTM with a hidden size of $256$ as $\mathcal{A}$.
Furthermore, the decoders $\theta^{\mathcal{M}}$, ${\mathcal{R}}^{\mathcal{M}}$, $\theta^{\mathcal{D}}$, ${\mathcal{R}}^{\mathcal{D}}$, and ${\mathcal{C}}^{\mathcal{D}}$ are structured with a fully connected (FC) of size $256$ layer followed by ReLU activation, five consecutive ResNet blocks~\cite{ResNet} of $256$, and another FC layer.
To ensure that the angle values fall within the range of $[-\pi,\pi]$, we employ $\tanh$ on the output of the decoders $\theta^{\mathcal{M}}$ and $\theta^{\mathcal{D}}$ and multiply the results by $\pi$.
Since there is no continuous size of the tools, we consider a predefined radius range $r\in radius=\{0.025,0.05,0.075,0.1\}$ for milling tools and $r\in radius=\{0.01,0.02,0.03,0.04\}$ for drilling tools.
To allow the decoders ${\mathcal{R}}^{\mathcal{M}}$ and ${\mathcal{R}}^{\mathcal{D}}$ to differentially select among the tool $radius$, we employ softmax on the output of the decoders and multiply them by the vector $radius$.
On the other hand, the path decoder $\mathcal{P}^{\mathcal{M}}$ has two branches, one to generate the depth of tool penetration $c_z$ using features obtained from $\mathcal{A}$, while the second branch, by the same features, receives the time step $t\in\{0,0.01,0.02,\dots,0.99\}$ as input to generate $(c_x,c_y)$.
Both branches have network architectures similar to those of the other decoders.
In all experiments, we set the resolution of voxels $\mathcal{V}_s$ as $64\times64\times64$, pick a large scaling factor $w=1000$, and limit the number of $20$ for both types of milling and drilling operations.
All experiments are carried out with PyTorch 1.12.0 and Quadro RTX 8000 GPUs.
We will publish our CNC-Net implementation.

\section{Experiments}
This section provides comprehensive information on training and evaluation datasets, training configurations, and a deeper analysis of our proposed method, CNC-Net.

\subsection{Dataset}
\noindent
\textbf{ABC.} 
The ABC~\cite{Abc} dataset comprises one million 3D Computer-Aided Design~(CAD) models, particularly in manufacturing CAD objects, which serves as a valuable resource for developing geometric deep-learning methods and applications.
To pre-train our method, we follow CAPRI-Net~\cite{CAPRI-Net}, sampling $5,000$ normalized single-part CAD objects.
Given the time-consuming nature of fine-tuning for each shape, we randomly sample $50$ shapes from $1,000$ test samples for fine-tuning and evaluation.

\noindent
\textbf{ShapeNet.} 
Additionally, we utilize a broader spectrum of objects sourced from the ShapeNet Core~(V1)~\cite{Shapenet} dataset. 
In our training and evaluation, we use the watertight shapes derived from ONet~\cite{ONet}.
Following CAPRI-Net~\cite{CAPRI-Net}, we subsample $35k$ shapes across $13$ categories for pre-training, and further, we randomly select $10$ shapes from the test set of each category for fine-tuning and evaluation.

\subsection{Evaluation metrics}
Quantitative evaluations encompass widely used metrics, including volume-based metrics Intersection over Union~(IoU)~\cite{IoU} and F1~\cite{ExtrudeNet}, and surface-based metrics symmetric Chamfer Distance~(CD)~\cite{CD} and Normal Consistency~(NC)~\cite{ECD}.
To measure IoU and F1 for all methods, we voxelize the box $[-0.5,0.5]^3\subset\mathbb{R}^3$ into $256^3$ voxels and evaluate their occupancies from the reconstructed meshes.
For CD and NC measurements, following CAPRI-Net~\cite{CAPRI-Net}, we uniformly sample $8k$ points on the surface of each object, where all CD values are multiplied by $1,000$.

%
%

\subsection{Training and evaluation}
While pre-training provides foundational knowledge and learning generic features, fine-tuning adapts the model to specific samples, enhancing its effectiveness.
Initially, we pre-train our model with the self-supervised objectives defined in~\cref{sec:losses}.
This pre-training phase involves $100$ epochs on training samples from ABC~\cite{Abc} and ShapeNet~\cite{Shapenet} datasets, taking $0.5$ and $1.5$ hours per epoch, respectively. 
Later, given the fully self-supervised nature of our method, proceed to fine-tune the pre-trained model, following prior approaches~\cite{CSG-Stump, ExtrudeNet, CAPRI-Net, SECAD-Net} for $12,000$ iterations on each test sample individually, which takes around $30$ minutes per sample.
We utilize ADAM~\cite{ADAM} optimizer with a learning rate of $1\times10^{-4}$ for both pre-training and fine-tuning experiments on both ABC~\cite{Abc} and ShapeNet~\cite{Shapenet} datasets.

We conduct various quantitative and qualitative experiments on both ABC~\cite{Abc} and ShapeNet~\cite{Shapenet} datasets to compare our reconstruction performance in contrast to state-of-the-art~(SOTA) 3D CAD reconstruction methods including CSG-Stump~\cite{CSG-Stump}, ExtrudeNet~\cite{ExtrudeNet}, CAPRI-Net~\cite{CAPRI-Net}, and SECAD-Net~\cite{SECAD-Net}.
For fair comparisons, we use the existing pre-trained models of CAPRI-Net~\cite{CAPRI-Net} and SECAD-Net~\cite{SECAD-Net}, while we pre-train CSG-Stump~\cite{CSG-Stump} and ExtrudeNet~\cite{ExtrudeNet} using their official implementations.
Furthermore, the fine-tuning process for all methods involves the same number of iterations.

\subsubsection{Results on ABC dataset}
The results shown in \cref{tab:quant_abc} on ABC~\cite{Abc} dataset illustrate the superior performance of our self-supervised method in accurately reproducing target 3D CAD models by simulating CNC machining operations. 
%
In particular, compared to other 3D CAD reconstruction techniques, our CNC-Net method can improve the performance of CSG-Stump~\cite{CSG-Stump}, ExtrudeNet~\cite{ExtrudeNet}, CAPRI-Net~\cite{CAPRI-Net}, and SECAD-Net~\cite{SECAD-Net} by $4.7\%$, $7.2\%$, $7.3\%$, and $6.2\%$ of IoU metric, respectively.
We observe an inferior performance of our method compared to other methods when considering surface-based metrics CD and NC. 
This discrepancy arises because prior methods generate shapes by combining several smooth primitives, while our approach carves a cube to approximate the shape. 
Consequently, our approach struggles to achieve the same level of smoothness in the reproduced shapes.
%
%
%
%
\begin{table}[h]
    \centering
    \small
    \resizebox{\linewidth}{!}{\begin{tabular}{lcccc}
        \toprule
        \textbf{Method} & 
        \textbf{IoU}$\uparrow$ & 
        \textbf{F1}$\uparrow$ &
        \textbf{CD}$\downarrow$& 
        \textbf{NC}$\uparrow$ 
          \\
        \midrule
        CSG-Stump~\cite{CSG-Stump} &  {0.787} & {0.879} & {0.428} & {0.884} \\
        ExtrudeNet~\cite{ExtrudeNet} &  {0.769} & {0.875} & {0.505} & {0.871} \\
        CAPRI-Net~\cite{CAPRI-Net} &  {0.768} & {0.866} & \textbf{0.312} & \textbf{0.914} \\        
        SECAD-Net~\cite{SECAD-Net} &  {0.776} & {0.867}  & {0.398} & {0.900}\\
        \midrule
        \textbf{CNC-Net~(Ours)} & \textbf{0.824} & \textbf{0.901} & {0.509} & {0.893}  \\
        \bottomrule
    \end{tabular}}
    \caption{
        \textbf{Quantitative results on ABC~\cite{Abc} dataset.}
    }\label{tab:quant_abc}
    \vspace{-4mm}
\end{table} 

Moreover, we provide a visual comparison of our results with those of CSG-Stump~\cite{CSG-Stump}, ExtrudeNet~\cite{ExtrudeNet}, CAPRI-Net~\cite{CAPRI-Net}, and SECAD-Net~\cite{SECAD-Net} on ABC dataset~\cite{Abc} depicted in \cref{fig:qual_abc}.
For equitable comparison, all reconstructed CAD models are visualized using marching cubes~(MC) at a resolution of $256$.
Qualitative comparisons highlight the superiority of our CNC-Net in faithfully reproducing the overall shape of the target CAD models while exhibiting exceptional precision in preserving more local details, setting it apart from other methods.
Specifically, taking advantage of designed carving and rotation operations, our method adeptly generates holes via drilling, a capability not achieved by CSG-Stump~\cite{CSG-Stump} and ExtrudeNet~\cite{ExtrudeNet}.
Furthermore, CNC-Net has the advantage of preserving parts that CAPRI-Net~\cite{CAPRI-Net} and SECAD-Net~\cite{SECAD-Net} might damage. 
This aspect is beneficial as our reproduced shapes can be further refined through post-processing.

Therefore, the precise reproduction from 3D CAD models achieved by our CNC-Net method, as evident in quantitative and qualitative experiments, emphasizes its practical applicability in shaping desired objects from raw materials by learning CNC machine operations.
We provide more qualitative results in our supplementary material.

\subsubsection{Results on ShapeNet dataset}
\begin{figure}[h]
     \centering
     \begin{subfigure}[b]{0.15\linewidth}
        \centering
        \includegraphics[trim={0 1cm 0 0},clip,width=\linewidth, page=1]{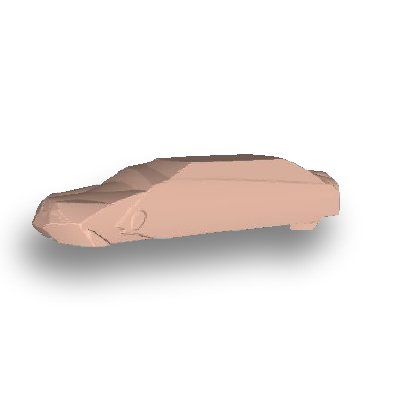}
     \end{subfigure}
     \hfill
     \begin{subfigure}[b]{0.15\linewidth}
        \centering
        \includegraphics[trim={0 1cm 0 0},clip,width=\linewidth, page=2]{figures/ShapeNet_car.pdf}
     \end{subfigure}
     \hfill
     \begin{subfigure}[b]{0.15\linewidth}
        \centering
        \includegraphics[trim={0 1cm 0 0},clip,width=\linewidth, page=3]{figures/ShapeNet_car.pdf}
     \end{subfigure}
     \hfill
     \begin{subfigure}[b]{0.15\linewidth}
        \centering
        \includegraphics[trim={0 1cm 0 0},clip,width=\linewidth, page=4]{figures/ShapeNet_car.pdf}
     \end{subfigure}
     \hfill
     \begin{subfigure}[b]{0.15\linewidth}
        \centering
        \includegraphics[trim={0 1cm 0 0},clip,width=\linewidth, page=5]{figures/ShapeNet_car.pdf}
     \end{subfigure}
     \hfill
     \begin{subfigure}[b]{0.15\linewidth}
        \centering
        \includegraphics[trim={0 1cm 0 0},clip,width=\linewidth, page=6]{figures/ShapeNet_car.pdf}
     \end{subfigure}
     
     \begin{subfigure}[b]{0.15\linewidth}
        \centering
        \includegraphics[trim={0 0 0 1cm},clip,width=\linewidth, page=1]{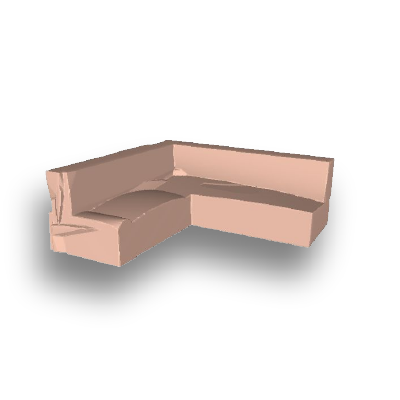}
     \end{subfigure}
     \hfill
     \begin{subfigure}[b]{0.15\linewidth}
        \centering
        \includegraphics[trim={0 0 0 1cm},clip,width=\linewidth, page=2]{figures/ShapeNet_sofa.pdf}
     \end{subfigure}
     \hfill
     \begin{subfigure}[b]{0.15\linewidth}
        \centering
        \includegraphics[trim={0 0 0 1cm},clip,width=\linewidth, page=3]{figures/ShapeNet_sofa.pdf}
     \end{subfigure}
     \hfill
     \begin{subfigure}[b]{0.15\linewidth}
        \centering
        \includegraphics[trim={0 0 0 1cm},clip,width=\linewidth, page=4]{figures/ShapeNet_sofa.pdf}
     \end{subfigure}
     \hfill
     \begin{subfigure}[b]{0.15\linewidth}
        \centering
        \includegraphics[trim={0 0 0 1cm},clip,width=\linewidth, page=5]{figures/ShapeNet_sofa.pdf}
     \end{subfigure}
     \hfill
     \begin{subfigure}[b]{0.15\linewidth}
        \centering
        \includegraphics[trim={0 0 0 1cm},clip,width=\linewidth, page=6]{figures/ShapeNet_sofa.pdf}
     \end{subfigure}

     \begin{subfigure}[b]{0.15\linewidth}
        \centering
        \includegraphics[trim={0 0 0 2cm},clip,width=\linewidth, page=1]{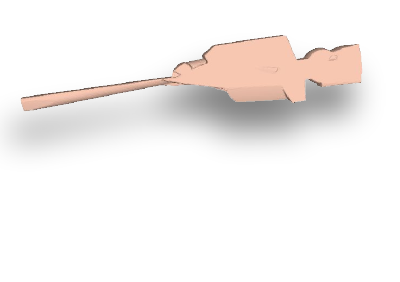}
     \end{subfigure}
     \hfill
     \begin{subfigure}[b]{0.15\linewidth}
        \centering
        \includegraphics[trim={0 0 0 2cm},clip,width=\linewidth, page=2]{figures/ShapeNet_gun.pdf}
     \end{subfigure}
     \hfill
     \begin{subfigure}[b]{0.15\linewidth}
        \centering
        \includegraphics[trim={0 0 0 2cm},clip,width=\linewidth, page=3]{figures/ShapeNet_gun.pdf}
     \end{subfigure}
     \hfill
     \begin{subfigure}[b]{0.15\linewidth}
        \centering
        \includegraphics[trim={0 0 0 2cm},clip,width=\linewidth, page=4]{figures/ShapeNet_gun.pdf}
     \end{subfigure}
     \hfill
     \begin{subfigure}[b]{0.15\linewidth}
        \centering
        \includegraphics[trim={0 0 0 2cm},clip,width=\linewidth, page=5]{figures/ShapeNet_gun.pdf}
     \end{subfigure}
     \hfill
     \begin{subfigure}[b]{0.15\linewidth}
        \centering
        \includegraphics[trim={0 0 0 2cm},clip,width=\linewidth, page=6]{figures/ShapeNet_gun.pdf}
     \end{subfigure}
     
     \begin{subfigure}[b]{0.15\linewidth}
        \centering
        \includegraphics[trim={0 0 0 2.5cm},clip,width=\linewidth, page=1]{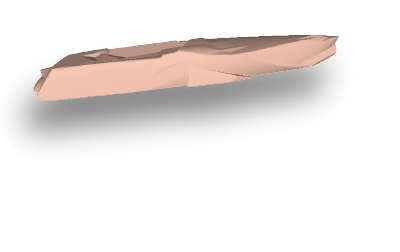}
        \caption*{\tiny CSG-Stump~\cite{CSG-Stump}}
     \end{subfigure}
     \hfill
    \begin{subfigure}[b]{0.15\linewidth}
        \centering
        \includegraphics[trim={0 0 0 2.5cm},clip,width=\linewidth, page=1]{figures/ShapeNet_ship.pdf}
        \caption*{\tiny ExtrudeNet~\cite{ExtrudeNet}}
     \end{subfigure}
     \hfill
     \begin{subfigure}[b]{0.15\linewidth}
        \centering
        \includegraphics[trim={0 0 0 2.5cm},clip,width=\linewidth, page=3]{figures/ShapeNet_ship.pdf}
        \caption*{\tiny CAPRI-Net~\cite{CAPRI-Net}}
     \end{subfigure}
    \hfill
     \begin{subfigure}[b]{0.15\linewidth}
        \centering
        \includegraphics[trim={0 0 0 2.5cm},clip,width=\linewidth, page=4]{figures/ShapeNet_ship.pdf}
        \caption*{\tiny SECAD-Net~\cite{SECAD-Net}}
     \end{subfigure}
     \hfill
     \begin{subfigure}[b]{0.15\linewidth}
        \centering
        \includegraphics[trim={0 0 0 2.5cm},clip,width=\linewidth, page=5]{figures/ShapeNet_ship.pdf}
        \caption*{\tiny CNC-Net~(Ours)}
     \end{subfigure}
     \hfill
     \begin{subfigure}[b]{0.15\linewidth}
        \centering
        \includegraphics[trim={0 0 0 2.5cm},clip,width=\linewidth, page=6]{figures/ShapeNet_ship.pdf}
        \caption*{\tiny Target~(GT)}
     \end{subfigure}
        \caption{
        \textbf{Qualitative results on ShapeNet~\cite{Shapenet} dataset.} 
        All reproduced shapes are visualized using marching cubes~(MC) with $256$ resolution. The \nth{1} to \nth{4} rows visualize the results of sampled shapes from the car, sofa, rifle, and vessel categories, respectively.}
        \label{fig:qual_shapenet}
        \vspace{-4mm}
\end{figure}

The quantitative and qualitative comparisons of our method over SOTA 3D CAD reconstruction methods~\cite{CSG-Stump, ExtrudeNet, CAPRI-Net, SECAD-Net} on the ShapeNet~\cite{Shapenet} dataset are shown in \cref{tab:quant_shapenet} and \cref{fig:qual_shapenet}, respectively.
%
%
%
Although the objects in this dataset are generally unions of object parts, \eg, chairs composed of legs, back, seat bottomland, etc., and cannot be easily processed using generic CNC machines, our method still demonstrates superior performance in terms of IoU and F1 metrics.
As our method carves the shapes, it performs inferior compared to methods that construct shapes through unions in terms of surface-based metrics CD and NC.
We provide more visual results in our supplementary material.

\begin{table}[h]
    \centering
    \small
    \resizebox{\linewidth}{!}{\begin{tabular}{lcccc}
        \toprule
        \textbf{Method} & 
        \textbf{IoU}$\uparrow$ & 
        \textbf{F1}$\downarrow$& 
        \textbf{CD}$\downarrow$& 
        \textbf{NC}$\uparrow$   \\
        \midrule
        CSG-Stump~\cite{CSG-Stump} &  {0.697} & {0.827} & {0.521} & {0.866} \\        
        ExtrudeNet~\cite{ExtrudeNet} & {0.607} & {0.773} & {0.918} & {0.844} \\
        CAPRI-Net~\cite{CAPRI-Net} &  {0.700}  & {0.824} & \textbf{0.447} & \textbf{0.895} \\
        SECAD-Net~\cite{SECAD-Net} &  {0.650} & {0.784} & {2.405} & {0.852} \\
        \midrule
        \textbf{CNC-Net~(Ours)} & \textbf{0.740} & \textbf{0.850} & {1.562} & {0.863}  \\
        \bottomrule
    \end{tabular}}
    \caption{
        \textbf{Quantitative results on ShapeNet~\cite{Shapenet} dataset.}
    }\label{tab:quant_shapenet}
    \vspace{-4mm}
\end{table} 

\subsection{Ablation study}
We further comprehensively analyze our proposed CNC-Net through a series of ablation studies. 
In these studies, we illustrate the learned milling paths and the zero-shot learning capability, explore the effect of various loss functions, and investigate the impact of each operation.

\subsubsection{Milling paths}
\begin{figure}[h]
\centering
\small
    \begin{subfigure}[b]{\linewidth}
    \begin{tabular}{cccccc}
         \includegraphics[page=1,width=.23\linewidth]{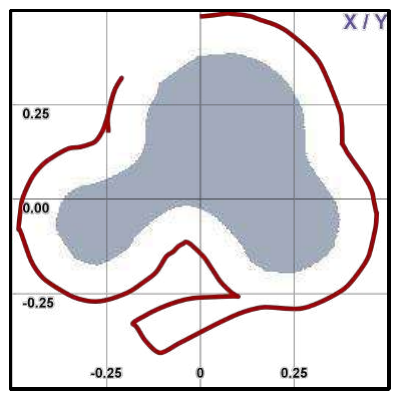}
         &\hspace{-4mm}
         \includegraphics[page=2,width=.23\linewidth]{figures/Path_22.pdf}
         &\hspace{-4mm}
         \includegraphics[page=3,width=.23\linewidth]{figures/Path_22.pdf}
         &\hspace{-4mm}
         \includegraphics[page=4,width=.23\linewidth]{figures/Path_22.pdf}
\\
         \includegraphics[page=1,width=.23\linewidth]{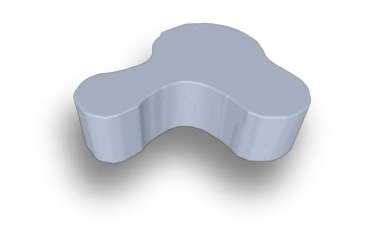}
         &\hspace{-4mm}
         \includegraphics[page=2,width=.23\linewidth]{figures/Path_22_CAD.pdf}
         &\hspace{-4mm}
         \includegraphics[page=3,width=.23\linewidth]{figures/Path_22_CAD.pdf}
         &\hspace{-4mm}
         \includegraphics[page=4,width=.23\linewidth]{figures/Path_22_CAD.pdf}
\\
         \includegraphics[page=1,width=.23\linewidth]{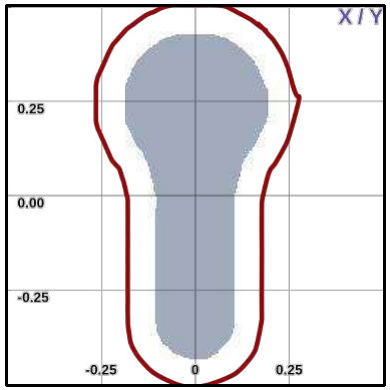}
         &\hspace{-4mm}
         \includegraphics[page=2,width=.23\linewidth]{figures/Path_707.pdf}
         &\hspace{-4mm}
         \includegraphics[page=3,width=.23\linewidth]{figures/Path_707.pdf}
         &\hspace{-4mm}
         \includegraphics[page=4,width=.23\linewidth]{figures/Path_707.pdf}
\\
         \includegraphics[page=1,width=.23\linewidth]{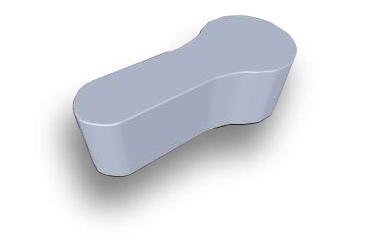}
         &\hspace{-4mm}
         \includegraphics[page=2,width=.23\linewidth]{figures/Path_707_CAD.pdf}
         &\hspace{-4mm}
         \includegraphics[page=3,width=.23\linewidth]{figures/Path_707_CAD.pdf}
         &\hspace{-4mm}
         \includegraphics[page=4,width=.23\linewidth]{figures/Path_707_CAD.pdf}
\\
         {\tiny $\mathcal{P}_{1},\mathcal{S}_{1}$} 
         &\hspace{-4mm} {\tiny $\mathcal{P}_{2},\mathcal{S}_{2}$} 
         &\hspace{-4mm} {\tiny $\mathcal{P}_{3},\mathcal{S}_{3}$}
         &\hspace{-4mm} {\tiny $\mathcal{P}_{4},\mathcal{S}_{4}$}
         \end{tabular}
     \end{subfigure}
    \caption{
        \textbf{Ablation study for milling paths.}
        The \nth{1} and \nth{3} rows display the path $\mathcal{P}$ in steps $s=1,\dots,4$ from the top view.
        The \nth{2} and \nth{4} rows depict the reproduced shapes in each step.
    }
    \label{fig:ablations_path}
    \vspace{-4mm}
\end{figure}

We visualize the learned path  $\mathcal{P}_s$ for steps $s=1,\dots,4$ and the generated shape $\mathcal{S}_s$ after applying the milling operation in \cref{fig:ablations_path}.
The results indicate that the decoder $\mathcal{P}^{\mathcal{M}}$ generates a path that outlines the boundary of the target object to form its overall shape in the first step and shorter paths in later steps to reach the fine details.

\subsubsection{Zero-shot learning on a single shape}
Based on the fully self-supervised nature of our method, we show the advantage of our CNC-Net to train in a zero-shot manner for each sample individually, eliminating the need for a large-scale training dataset.
%
Consequently, we train the model individually for each $50$ and $130$ test sample from the ABC~\cite{Abc} and ShapeNet~\cite{Shapenet} datasets, respectively.
Comparisons between zero-shot and fine-tuned results indicate that our method can reproduce target shapes without significant degradation performance, as shown in \cref{tab:ablation_zeroshot}.
In subsequent ablation studies, we show the results for the models trained in a zero-shot manner.
\begin{table}[h]
\centering
\small
    \resizebox{\linewidth}{!}{\begin{tabular}{lccccccccc}
        \toprule
        \multirow{2}{*}{\textbf{Training}} &&  \multicolumn{4}{c}{ABC~\cite{Abc}} & \multicolumn{4}{c}{ShapeNet~\cite{Shapenet}}
        \\
        \cline{2-5}\cline{7-10}&
        \textbf{IoU$\uparrow$} & 
        \textbf{F1$\uparrow$} &        
        \textbf{CD$\downarrow$} & 
        \textbf{NC$\uparrow$} 
        &&
        \textbf{IoU$\uparrow$} & 
        \textbf{F1$\uparrow$} &        
        \textbf{CD$\downarrow$} & 
        \textbf{NC$\uparrow$} 
        \\
        \hline
         Fine-tuning & \textbf{0.824} & \textbf{0.901} & \textbf{0.509} & \textbf{0.893} && \textbf{0.740} & \textbf{0.850} & \textbf{1.562} & \textbf{0.863} \\
         Zero-shot & {0.780} & {0.878} & {1.127} & {0.870} && {0.698} & {0.812} & {2.155} & {0.845} \\
        \hline
    \end{tabular}}
    \caption{
        \textbf{Ablation study for zero-shot training.}
    }
    \label{tab:ablation_zeroshot}
    \vspace{-4mm}
\end{table}

\subsubsection{Effect of losses}\label{sec:abl_loss}
We evaluate the effect of our defined loss functions in \cref{sec:losses} on the test samples of ABC~\cite{Abc} dataset detailed in \cref{tab:ablation_loss}.
The results demonstrate that our method can be effectively trained using only the essential loss functions $\mathcal{L}^{\mathcal{M}}$ and $\mathcal{L}^{\mathcal{D}}$, while the collaboration of
$\mathcal{L}^{\mathcal{S}}$ and $\mathcal{L}^{\mathcal{C}}$ serves as a guiding factor to improve the reconstruction performance.
\begin{table}[h]
\centering
\small
    \resizebox{\linewidth}{!}{\begin{tabular}{lccccccc}
        \toprule
        $\mathcal{L}^{\mathcal{M}}$&
        $\mathcal{L}^{\mathcal{D}}$&
        $\mathcal{L}^{\mathcal{S}}$&
        $\mathcal{L}^{\mathcal{C}}$ &
        \textbf{IoU$\uparrow$} & 
        \textbf{F1$\uparrow$} &
        \textbf{CD$\downarrow$} & 
        \textbf{NC$\uparrow$} 
        \\
        \hline
         $\cmark[blue, scale=0.5]$ & $\cmark[blue, scale=0.5]$ & $\crossmark[red, scale=0.5]$ & $\crossmark[red, scale=0.5]$ & {0.721} & {0.845} & {5.202} & {0.837} \\
         $\cmark[blue, scale=0.5]$ & $\cmark[blue, scale=0.5]$ & $\cmark[blue, scale=0.5]$ & $\crossmark[red, scale=0.5]$ & {0.686} & {0.820} & {5.138} & {0.831} \\
         $\cmark[blue, scale=0.5]$ & $\cmark[blue, scale=0.5]$ & $\crossmark[red, scale=0.5]$ & $\cmark[blue, scale=0.5]$ & {0.755} & {0.847} & {7.184} & {0.859} \\
         \hline
         $\cmark[blue, scale=0.5]$ & $\cmark[blue, scale=0.5]$ & $\cmark[blue, scale=0.5]$ &$\cmark[blue, scale=0.5]$ & \textbf{0.780} & \textbf{0.878} & \textbf{1.127} & \textbf{0.870}\\
        \hline
    \end{tabular}}
    \caption{
        \textbf{Ablation study for loss functions.}
    }
    \label{tab:ablation_loss}
    \vspace{-4mm}
\end{table}

\subsubsection{Effect of operations}\label{sec:abl_eff_op}
We further explore the individual impact of milling $\mathcal{O}^{\mathcal{M}}$, drilling $\mathcal{O}^{\mathcal{D}}$, and rotation $\mathcal{R}ot$ operations by excluding each one during both the training and testing phases.
The results in \cref{tab:ablation_operation} on the ABC dataset~\cite{Abc} underscore the significant role of milling and rotation operations in shaping target objects. 
In contrast, drilling slightly contributes to more precise shape reconstruction.
We observe that removing the drilling operation results in almost twice the value for the CD metric. 
The reason is that to generate a hole in the shape, the milling operation has to function as a drilling process, resulting in a non-smooth shape.
\begin{table}[h]
\centering
\small
    \resizebox{\linewidth}{!}{\begin{tabular}{lcccccc}
        \toprule
        $\mathcal{O}^{\mathcal{M}}$&
        $\mathcal{O}^{\mathcal{D}}$&
        $\mathcal{R}ot$&
        \textbf{IoU$\uparrow$} & 
        \textbf{F1$\uparrow$} &        
        \textbf{CD$\downarrow$} & 
        \textbf{NC$\uparrow$} 
        \\
        \hline
        $\crossmark[red, scale=0.5]$  & $\cmark[blue, scale=0.5]$ & $\cmark[blue, scale=0.5]$ & {0.460} & {0.630} & {8.315} & {0.701} \\
         $\cmark[blue, scale=0.5]$ & $\cmark[blue, scale=0.5]$ & $\crossmark[red, scale=0.5]$ & {0.525} & {0.677} & {10.826} & {0.738} \\
         $\cmark[blue, scale=0.5]$ & $\crossmark[red, scale=0.5]$ & $\cmark[blue, scale=0.5]$ & {0.776} & {0.876} & {2.098} & {0.869} \\
         \hline
         $\cmark[blue, scale=0.5]$ & $\cmark[blue, scale=0.5]$ & $\cmark[blue, scale=0.5]$ & \textbf{0.780} & \textbf{0.878} & \textbf{1.127} & \textbf{0.870}\\
        \hline
    \end{tabular}}
    \caption{
        \textbf{Ablation study for operations.}
    }
    \label{tab:ablation_operation}
    \vspace{-4mm}
\end{table}

\section{Conclusion}
We propose CNC-Net, a novel self-supervised DNN-based framework designed to simulate a generic CNC machine.
CNC-Net provides a sequence of learnable modeled manufacturing operations with implicit representation to construct desired objects from raw materials. 
Our quantitative and qualitative reconstruction results demonstrate the superior performance of our method compared to the most state-of-the-art 3D CAD reconstruction techniques.  
\paragraph{Limitation and future work.}
One remaining limitation is that no dataset contains the sequential operations required to serve as a reference for our learned milling and drilling operations.
This absence makes it challenging to assess the efficiency of our method and determine whether it represents an optimal solution.
On the other hand, finding multi-stage execution for CNC machines is an NP-hard problem, so it is not time-efficient and requires expensive specialized human labor.
In future work, we plan to enhance our method by incorporating a decision-making process to select the appropriate type of operation at each step.
Although cylindrical tools are common, it is important to note that tool shapes are diverse and designed for specific machining operations and applications.
Our future endeavors include exploring tools with various shapes, such as broaches, gear, and fly cutters, and incorporating a broader range of operations such as grinding, turning, and lapping.
Additionally, our current focus on the input raw material as a bounding box can be expanded to encompass shapes with various geometry in future investigations.

\small
\bibliographystyle{ieee_fullname}
\bibliography{egbib}

\end{document}


\maketitle

\section{Inferior performance with CD and NC}
\begin{figure*}[b]
    \centering
	\captionsetup[subfloat]{labelformat=empty,aboveskip=1pt}
		\setlength\tabcolsep{0.0cm}
		\begin{tabular}[b]{c c c c}
			\subfloat[{\centering  CSG-Stump}]{
				\includegraphics[width=0.16\linewidth, page=1]
				{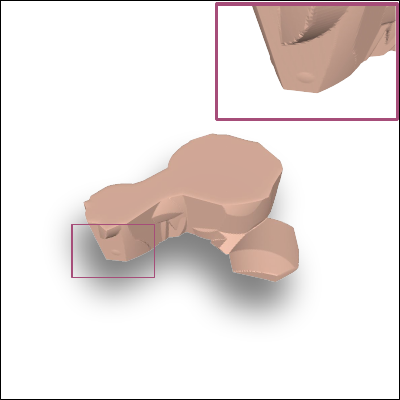}} %
            \fbox{\parbox{0.05\textwidth}{
            {\vspace{-40mm}\scriptsize
            IoU=0.882}\vspace{2mm}\\ {\scriptsize CD=0.322}\vspace{2mm}\\ {\scriptsize NC=0.913}
            }}
            \hspace{2mm}
			\subfloat[\centering  CAPRI-Net  ]{
			    \includegraphics[width=0.16\linewidth, page=2]
				{figures/smooth_2.pdf}}
            \fbox{\parbox{0.05\textwidth}{
            {\vspace{-40mm}\scriptsize
            IoU=0.901}\vspace{2mm}\\ {\scriptsize CD=0.166}\vspace{2mm}\\ {\scriptsize NC=0.926}
            }}
            \hspace{2mm}
			\subfloat[\centering  SECAD-Net]{
				\includegraphics[width=0.16\linewidth, page=3]
				{figures/smooth_2.pdf}} %
            \fbox{\parbox{0.05\textwidth}{
            {\vspace{-40mm}\scriptsize
            IoU=0.894}\vspace{2mm}\\ {\scriptsize \bf CD=0.159}\vspace{2mm}\\ {\scriptsize \bf NC=0.929}
            }}
            \hspace{2mm}
			\subfloat[\centering  CNC-Net]{
				\includegraphics[width=0.16\linewidth, page=4]
				{figures/smooth_2.pdf}}
            \fbox{\parbox{0.05\textwidth}{
            {\vspace{-40mm}\scriptsize
            {\bf IoU=0.930}}\vspace{2mm}\\ {\scriptsize { CD=0.329}}\vspace{2mm}\\ {\scriptsize NC=0.908}
            }}
	\end{tabular}
 
		\begin{tabular}[b]{c c c c}
			\subfloat[{\centering  CSG-Stump}]{
				\includegraphics[width=0.16\linewidth, page=1]
				{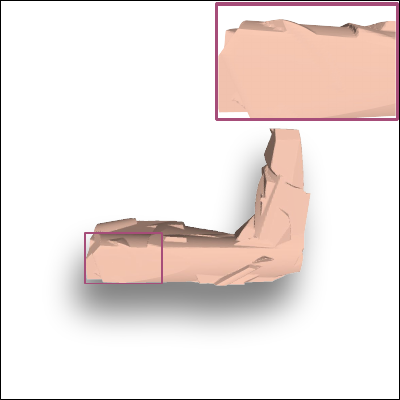}} %
            \fbox{\parbox{0.05\textwidth}{
            {\vspace{-40mm}\scriptsize
            IoU=0.619}\vspace{2mm}\\ {\scriptsize CD=0.805}\vspace{2mm}\\ {\scriptsize NC=0.750}
            }}
            \hspace{2mm}
			\subfloat[\centering  CAPRI-Net  ]{
			    \includegraphics[width=0.16\linewidth, page=2]
				{figures/smooth_3.pdf}}
            \fbox{\parbox{0.05\textwidth}{
            {\vspace{-40mm}\scriptsize
            IoU=0.240}\vspace{2mm}\\ {\scriptsize CD=0.668}\vspace{2mm}\\ {\bf \scriptsize NC=0.845}
            }}
            \hspace{2mm}
			\subfloat[\centering  SECAD-Net]{
				\includegraphics[width=0.16\linewidth, page=3]
				{figures/smooth_3.pdf}} %
            \fbox{\parbox{0.05\textwidth}{
            {\vspace{-40mm}\scriptsize
            IoU=0.132}\vspace{2mm}\\ {\scriptsize CD=0.464}\vspace{2mm}\\ {\scriptsize NC=0.783}
            }}
            \hspace{2mm}
			\subfloat[\centering  CNC-Net]{
				\includegraphics[width=0.16\linewidth, page=4]
				{figures/smooth_3.pdf}}
            \fbox{\parbox{0.05\textwidth}{
            {\vspace{-40mm}\scriptsize
            {\bf IoU=0.724}}\vspace{2mm}\\ {\scriptsize {\bf CD=0.350}}\vspace{2mm}\\ {\scriptsize NC=0.783}
            }}
	\end{tabular}

		\begin{tabular}[b]{c c c c}
			\subfloat[{\centering  CSG-Stump}]{
				\includegraphics[width=0.16\linewidth, page=1]
				{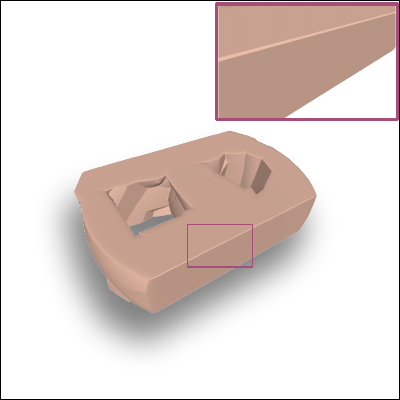}} %
            \fbox{\parbox{0.05\textwidth}{
            {\vspace{-40mm}\scriptsize
            IoU=0.880}\vspace{2mm}\\ {\scriptsize CD=0.223}\vspace{2mm}\\ {\scriptsize NC=0.922}
            }}
            \hspace{2mm}
			\subfloat[\centering  CAPRI-Net  ]{
			    \includegraphics[width=0.16\linewidth, page=2]
				{figures/smooth_1.pdf}}
            \fbox{\parbox{0.05\textwidth}{
            {\vspace{-40mm}\scriptsize
            IoU=0.853}\vspace{2mm}\\ {\scriptsize CD=0.235}\vspace{2mm}\\ {\scriptsize \bf NC=0.961}
            }}
            \hspace{2mm}
			\subfloat[\centering  SECAD-Net]{
				\includegraphics[width=0.16\linewidth, page=3]
				{figures/smooth_1.pdf}} %
            \fbox{\parbox{0.05\textwidth}{
            {\vspace{-40mm}\scriptsize
            IoU=0.882}\vspace{2mm}\\ {\scriptsize CD=0.205}\vspace{2mm}\\ {\scriptsize NC=0.958}
            }}
            \hspace{2mm}
			\subfloat[\centering  CNC-Net]{
				\includegraphics[width=0.16\linewidth, page=4]
				{figures/smooth_1.pdf}}
            \fbox{\parbox{0.05\textwidth}{
            {\vspace{-40mm}\scriptsize
            \bf IoU=0.908}\vspace{2mm}\\ {\scriptsize \bf CD=0.130}\vspace{2mm}\\ {\scriptsize NC=0.938}
            }}
	\end{tabular}

		\begin{tabular}[b]{c c c c}
			\subfloat[{\centering  CSG-Stump}]{
				\includegraphics[width=0.16\linewidth, page=1]
				{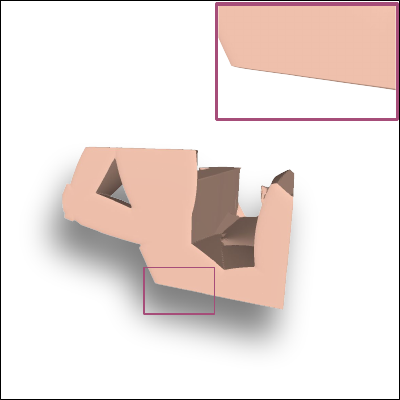}} %
            \fbox{\parbox{0.05\textwidth}{
            {\vspace{-40mm}\scriptsize
            IoU=0.807}\vspace{2mm}\\ {\scriptsize CD=0.353}\vspace{2mm}\\ {\scriptsize NC=0.851}
            }}
            \hspace{2mm}
			\subfloat[\centering  CAPRI-Net  ]{
			    \includegraphics[width=0.16\linewidth, page=2]
				{figures/smooth_4.pdf}}
            \fbox{\parbox{0.05\textwidth}{
            {\vspace{-40mm}\scriptsize
            IoU=0.854}\vspace{2mm}\\ {\scriptsize CD=0.341}\vspace{2mm}\\ {\scriptsize \bf NC=0.902}
            }}
            \hspace{2mm}
			\subfloat[\centering  SECAD-Net]{
				\includegraphics[width=0.16\linewidth, page=3]
				{figures/smooth_4.pdf}} %
            \fbox{\parbox{0.05\textwidth}{
            {\vspace{-40mm}\scriptsize
            IoU=0.800}\vspace{2mm}\\ {\scriptsize CD=0.693}\vspace{2mm}\\ {\scriptsize NC=0.857}
            }}
            \hspace{2mm}
			\subfloat[\centering  CNC-Net]{
				\includegraphics[width=0.16\linewidth, page=4]
				{figures/smooth_4.pdf}}
            \fbox{\parbox{0.05\textwidth}{
            {\vspace{-40mm}\scriptsize
            \bf IoU=0.879}\vspace{2mm}\\ {\scriptsize \bf CD=0.273}\vspace{2mm}\\ {\scriptsize NC=0.883}
            }}
	\end{tabular}
	\caption{\textbf{Comparison on the surface of the reconstructed shapes.}
	}
	\label{fig:surface}
\end{figure*}

In Sec.~4.3.1 of the main manuscript, we delve into the comparative analysis of the performance of our CNC-Net method against state-of-the-art~(SOTA) techniques~[\textcolor{cvprblue}{23}, \textcolor{cvprblue}{29}, \textcolor{cvprblue}{30}, \textcolor{cvprblue}{37}].
In this comparison, our method exhibits inferior results in surface-based metrics, including CD~[\textcolor{cvprblue}{25}] and NC~[\textcolor{cvprblue}{4}], while excelling in volume-based metrics when compared with such SOTA methods. 
This discrepancy arises from the fundamental difference in approach: prior methods construct shapes by combining several smooth primitives, whereas our approach approximates shapes by carving a cube.
To visually illustrate this distinction, we offer a closer examination of the surface in the CAD models produced by our CNC-Net method versus those generated by SOTA methods~[\textcolor{cvprblue}{23}, \textcolor{cvprblue}{29}, \textcolor{cvprblue}{37}] in \cref{fig:surface}.
The comparisons demonstrate that the results of our CNC-Net method are superior to those of these methods in forming the target shape to these methods~[\textcolor{cvprblue}{23}, \textcolor{cvprblue}{29}, \textcolor{cvprblue}{37}] with the IoU metric that is directly reflected in the visual appearance. 
However, the discrepancy in performance, as indicated by CD and NC metrics, is attributed to the non-smooth surfaces resulting from our approach of carving a cube to approximate shapes.

\definecolor{shadecolor}{rgb}{1,.8,.3}

\setlength{\fboxsep}{0pt}%
\setlength{\fboxrule}{0pt}%

\begin{figure*}[h]
\centering
\small
    \begin{subfigure}[b]{\linewidth}        
    \begin{tabular}{cccccc}
         \includegraphics[page=1,width=.11\linewidth]{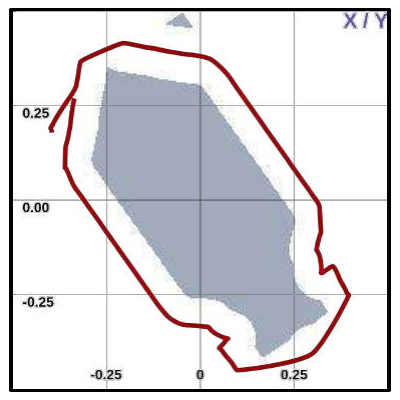}
         &\hspace{-4mm}
         \includegraphics[page=2,width=.11\linewidth]{figures/Path_190.pdf}
         &\hspace{-4mm}
         \includegraphics[page=3,width=.11\linewidth]{figures/Path_190.pdf}
         &\hspace{-4mm}
         \includegraphics[page=4,width=.11\linewidth]{figures/Path_190.pdf}
\\
         \includegraphics[page=1,width=.11\linewidth]{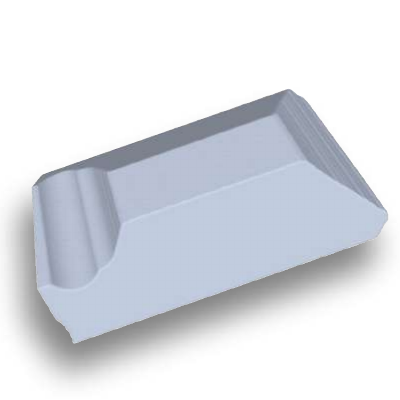}
         &\hspace{-4mm}
         \includegraphics[page=2,width=.11\linewidth]{figures/Path_190_CAD.pdf}
         &\hspace{-4mm}
         \includegraphics[page=3,width=.11\linewidth]{figures/Path_190_CAD.pdf}
         &\hspace{-4mm}
         \includegraphics[page=4,width=.11\linewidth]{figures/Path_190_CAD.pdf}
\\
         \includegraphics[page=1,width=.11\linewidth]{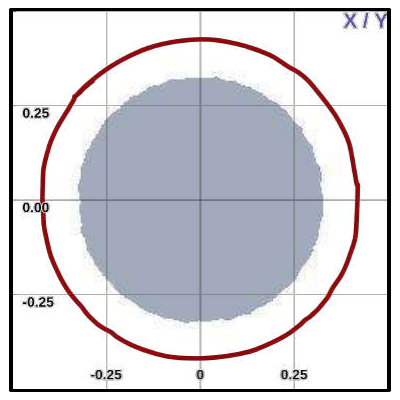}
         &\hspace{-4mm}
         \includegraphics[page=2,width=.11\linewidth]{figures/Path_214.pdf}
         &\hspace{-4mm}
         \includegraphics[page=3,width=.11\linewidth]{figures/Path_214.pdf}
         &\hspace{-4mm}
         \includegraphics[page=4,width=.11\linewidth]{figures/Path_214.pdf}
\\
         \includegraphics[page=1,width=.11\linewidth]{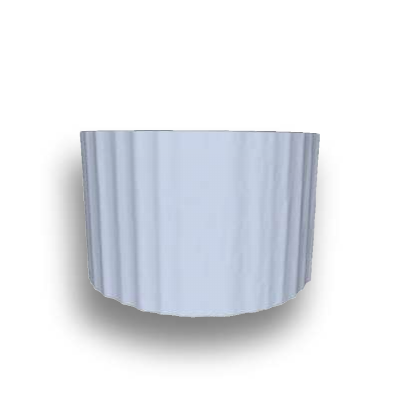}
         &\hspace{-4mm}
         \includegraphics[page=2,width=.11\linewidth]{figures/Path_214_CAD.pdf}
         &\hspace{-4mm}
         \includegraphics[page=3,width=.11\linewidth]{figures/Path_214_CAD.pdf}
         &\hspace{-4mm}
         \includegraphics[page=4,width=.11\linewidth]{figures/Path_214_CAD.pdf}
\\
         {\tiny $\mathcal{P}_{1},\mathcal{S}_{1}$} 
         &\hspace{-4mm} {\tiny $\mathcal{P}_{2},\mathcal{S}_{2}$} 
         &\hspace{-4mm} {\tiny $\mathcal{P}_{3},\mathcal{S}_{3}$}
         &\hspace{-4mm} {\tiny $\mathcal{P}_{4},\mathcal{S}_{4}$}
         \end{tabular}
\hfill
    \begin{tabular}{cccccc}
         \includegraphics[page=1,width=.11\linewidth]{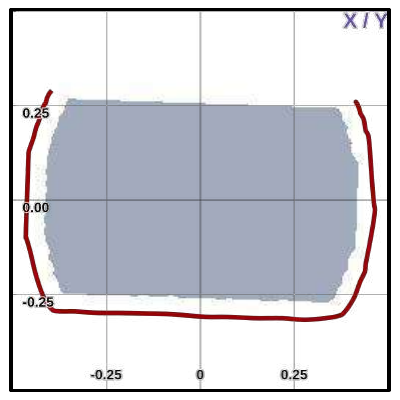}
         &\hspace{-4mm}
         \includegraphics[page=2,width=.11\linewidth]{figures/Path_729.pdf}
         &\hspace{-4mm}
         \includegraphics[page=3,width=.11\linewidth]{figures/Path_729.pdf}
         &\hspace{-4mm}
         \includegraphics[page=4,width=.11\linewidth]{figures/Path_729.pdf}
\\
         \includegraphics[page=1,width=.11\linewidth]{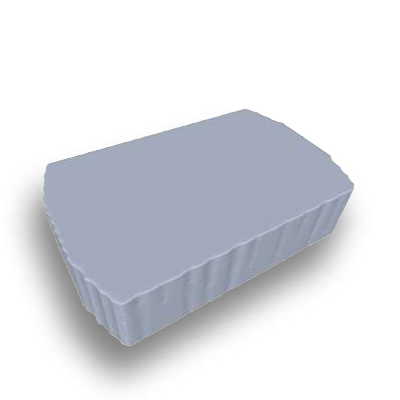}
         &\hspace{-4mm}
         \includegraphics[page=2,width=.11\linewidth]{figures/Path_729_CAD.pdf}
         &\hspace{-4mm}
         \includegraphics[page=3,width=.11\linewidth]{figures/Path_729_CAD.pdf}
         &\hspace{-4mm}
         \includegraphics[page=4,width=.11\linewidth]{figures/Path_729_CAD.pdf}
\\
         \includegraphics[page=1,width=.11\linewidth]{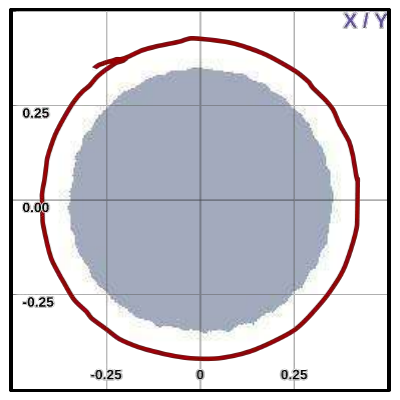}
         &\hspace{-4mm}
         \includegraphics[page=2,width=.11\linewidth]{figures/Path_440.pdf}
         &\hspace{-4mm}
         \includegraphics[page=3,width=.11\linewidth]{figures/Path_440.pdf}
         &\hspace{-4mm}
         \includegraphics[page=4,width=.11\linewidth]{figures/Path_440.pdf}
\\
         \includegraphics[page=1,width=.11\linewidth]{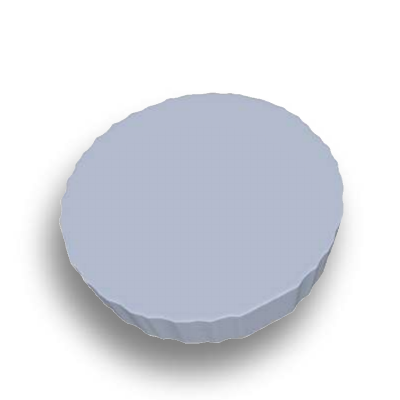}
         &\hspace{-4mm}
         \includegraphics[page=2,width=.11\linewidth]{figures/Path_440_CAD.pdf}
         &\hspace{-4mm}
         \includegraphics[page=3,width=.11\linewidth]{figures/Path_440_CAD.pdf}
         &\hspace{-4mm}
         \includegraphics[page=4,width=.11\linewidth]{figures/Path_440_CAD.pdf}
\\
         {\tiny $\mathcal{P}_{1},\mathcal{S}_{1}$} 
         &\hspace{-4mm} {\tiny $\mathcal{P}_{2},\mathcal{S}_{2}$} 
         &\hspace{-4mm} {\tiny $\mathcal{P}_{3},\mathcal{S}_{3}$}
         &\hspace{-4mm} {\tiny $\mathcal{P}_{4},\mathcal{S}_{4}$}
         \end{tabular}
     \end{subfigure}
    \caption{
        \textbf{Ablation study for milling paths.}
        The odd-numbered rows display the path $\mathcal{P}$ in steps $s=1,\dots,4$ from the top view.
        The even-numbered rows depict the reproduced shapes in each step.
    }
    \label{fig:path_supp}
    \vspace{-4mm}
\end{figure*}

\begin{figure*}[h]
     \centering
     \rotatebox[origin=lB]{90}{\bf Zero-shot}
     \hfill
    \begin{subfigure}[b]{0.12\linewidth}
        \centering
        \includegraphics[trim={0 0 0 0},clip,width=\linewidth, page=1]{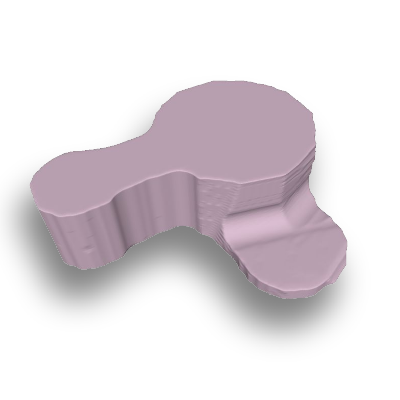}
     \end{subfigure}
     \hfill
     \begin{subfigure}[b]{0.12\linewidth}
        \centering
        \includegraphics[trim={0 0 0 0},clip,width=\linewidth, page=1]{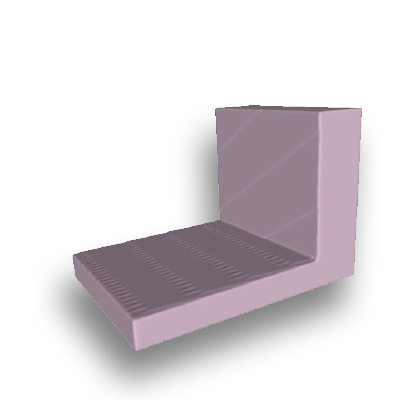}
     \end{subfigure}
     \hfill
     \begin{subfigure}[b]{0.12\linewidth}
        \centering
        \includegraphics[trim={0 0 0 0},clip,width=\linewidth, page=1]{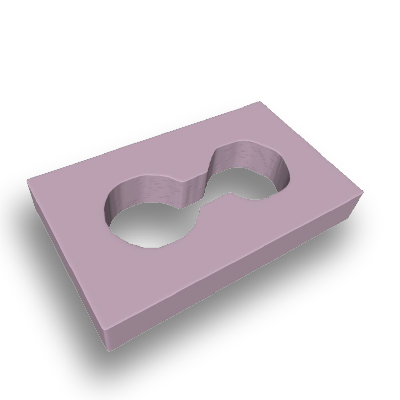}
     \end{subfigure}
     \hfill
     \begin{subfigure}[b]{0.12\linewidth}
        \centering
        \includegraphics[trim={0 0 0 0},clip,width=\linewidth, page=1]{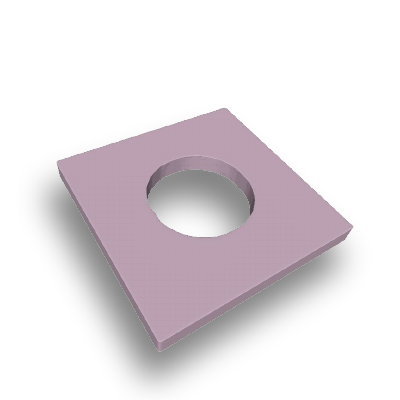}
     \end{subfigure}
     \hfill
     \begin{subfigure}[b]{0.12\linewidth}
        \centering
        \includegraphics[trim={0 0 0 0},clip,width=\linewidth, page=1]{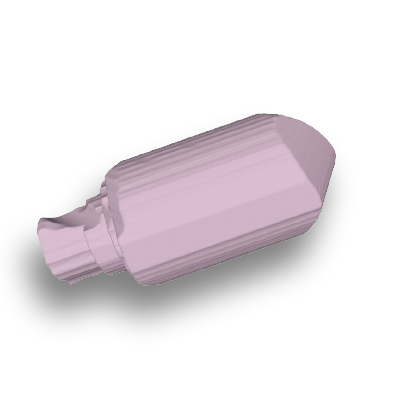}
     \end{subfigure}
     \hfill
     \begin{subfigure}[b]{0.12\linewidth}
        \centering
        \includegraphics[trim={0 0cm 0 0},clip,width=\linewidth, page=1]{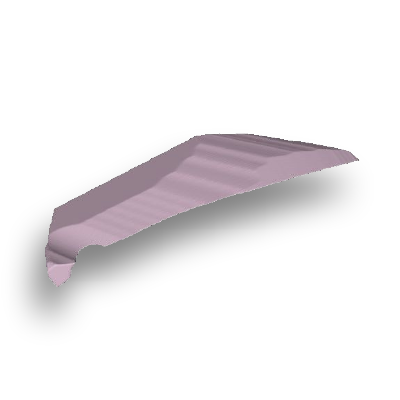}
     \end{subfigure}
     
     \rotatebox[origin=lB]{90}{\bf Fine-tuning}
     \hfill
     \begin{subfigure}[b]{0.12\linewidth}
        \centering
        \includegraphics[trim={0 0 0 0},clip,width=\linewidth, page=2]{figures/zero_1.pdf}
     \end{subfigure}
     \hfill
     \begin{subfigure}[b]{0.12\linewidth}
        \centering
        \includegraphics[trim={0 0 0 0},clip,width=\linewidth, page=2]{figures/zero_2.pdf}
     \end{subfigure}
     \hfill
     \begin{subfigure}[b]{0.12\linewidth}
        \centering
        \includegraphics[trim={0 0 0 0},clip,width=\linewidth, page=2]{figures/zero_3.pdf}
     \end{subfigure}
     \hfill
     \begin{subfigure}[b]{0.12\linewidth}
        \centering
        \includegraphics[trim={0 0 0 0},clip,width=\linewidth, page=2]{figures/zero_4.pdf}
     \end{subfigure}
     \hfill
     \begin{subfigure}[b]{0.12\linewidth}
        \centering
        \includegraphics[trim={0 0 0 0},clip,width=\linewidth, page=2]{figures/zero_5.pdf}
     \end{subfigure}
     \hfill
     \begin{subfigure}[b]{0.12\linewidth}
        \centering
        \includegraphics[trim={0 0 0 0},clip,width=\linewidth, page=2]{figures/zero_6.pdf}
     \end{subfigure}

     \rotatebox[origin=lB]{90}{\bf Target~(GT)}
     \hfill
     \begin{subfigure}[b]{0.12\linewidth}
        \centering
        \includegraphics[trim={0 0 0 0},clip,width=\linewidth, page=3]{figures/zero_1.pdf}
     \end{subfigure}
     \hfill
     \begin{subfigure}[b]{0.12\linewidth}
        \centering
        \includegraphics[trim={0 0 0 0},clip,width=\linewidth, page=3]{figures/zero_2.pdf}
     \end{subfigure}
     \hfill
     \begin{subfigure}[b]{0.12\linewidth}
        \centering
        \includegraphics[trim={0 0 0 0},clip,width=\linewidth, page=3]{figures/zero_3.pdf}
     \end{subfigure}
     \hfill
     \begin{subfigure}[b]{0.12\linewidth}
        \centering
        \includegraphics[trim={0 0 0 0},clip,width=\linewidth, page=3]{figures/zero_4.pdf}
     \end{subfigure}
     \hfill
     \begin{subfigure}[b]{0.12\linewidth}
        \centering
        \includegraphics[trim={0 0 0 0},clip,width=\linewidth, page=3]{figures/zero_5.pdf}
     \end{subfigure}
     \hfill
     \begin{subfigure}[b]{0.12\linewidth}
        \centering
        \includegraphics[trim={0 0 0 0},clip,width=\linewidth, page=3]{figures/zero_6.pdf}
     \end{subfigure}
        \caption{
        \textbf{Qualitative results for zero-shot Vs. fine-tuning} 
        }
        \label{fig:zero_vs_fine_supp}
        \vspace{-4mm}
\end{figure*}     

\begin{figure*}[!t]
     \centering
    \begin{subfigure}[b]{0.12\textwidth}
        \centering
        \includegraphics[trim={0 0 0 2cm},clip,width=\textwidth, page=1]{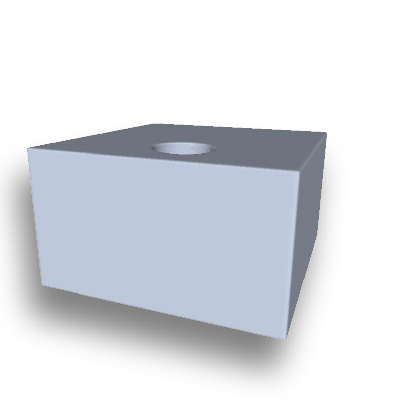}
     \end{subfigure}
     \hfill
     \begin{subfigure}[b]{0.12\textwidth}
        \centering
        \includegraphics[trim={0 0 0 2cm},clip,width=\textwidth, page=2]{figures/w_1.pdf}
     \end{subfigure}
     \hfill
     \begin{subfigure}[b]{0.12\textwidth}
        \centering
        \includegraphics[trim={0 0 0 2cm},clip,width=\textwidth, page=3]{figures/w_1.pdf}
     \end{subfigure}
     \hfill
     \begin{subfigure}[b]{0.12\textwidth}
        \centering
        \includegraphics[trim={0 0 0 2cm},clip,width=\textwidth, page=4]{figures/w_1.pdf}
     \end{subfigure}
     \hfill
     \begin{subfigure}[b]{0.12\textwidth}
        \centering
        \includegraphics[trim={0 0 0 2cm},clip,width=\textwidth, page=5]{figures/w_1.pdf}
     \end{subfigure}
     
     \begin{subfigure}[b]{0.12\textwidth}
        \centering
        \includegraphics[trim={0 0 0 0},clip,width=\textwidth, page=1]{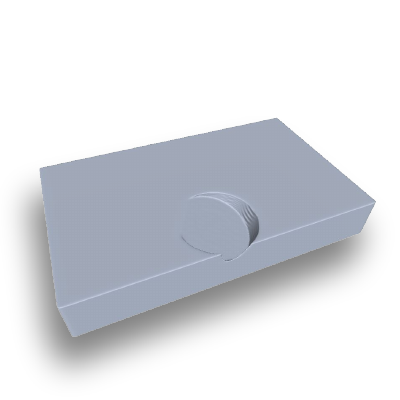}
        \caption*{Wo $\mathcal{O}^\mathcal{M}$}
     \end{subfigure}
    \hfill
     \begin{subfigure}[b]{0.12\textwidth}
        \centering
        \includegraphics[trim={0 0 0 0},clip,width=\textwidth, page=2]{figures/w_2.pdf}
        \caption*{Wo $\mathcal{O}^\mathcal{D}$}
     \end{subfigure}
     \hfill
     \begin{subfigure}[b]{0.12\textwidth}
        \centering
        \includegraphics[trim={0 0 0 0},clip,width=\textwidth, page=3]{figures/w_2.pdf}
        \caption*{Wo $\mathcal{R}ot$}
     \end{subfigure}
     \hfill
     \begin{subfigure}[b]{0.12\textwidth}
        \centering
        \includegraphics[trim={0 0 0 0},clip,width=\textwidth, page=4]{figures/w_2.pdf}
        \caption*{Zero-shot}
     \end{subfigure}
     \hfill
     \begin{subfigure}[b]{0.12\textwidth}
        \centering
        \includegraphics[trim={0 0 0 0},clip,width=\textwidth, page=5]{figures/w_2.pdf}
        \caption*{Target~(GT)}
     \end{subfigure}
        \caption{
        \textbf{Qualitative comparison of the results with different operations.} From the \nth{1} to the \nth{3} column, the milling, drilling, and rotation operation is eliminated, respectively.}
        \label{fig:wo}
        \vspace{-4mm}
\end{figure*}

\begin{figure*}[t]
     \centering
     \begin{subfigure}[b]{0.12\textwidth}
        \centering
        \includegraphics[trim={0 0.750cm 0 1.5cm},clip,width=\textwidth, page=1]{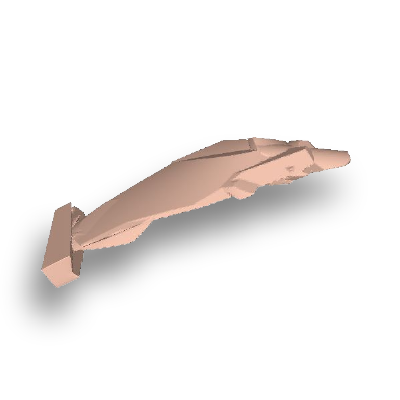}
     \end{subfigure}
     \hfill
     \begin{subfigure}[b]{0.12\textwidth}
        \centering
        \includegraphics[trim={0 0.750cm 0 1.5cm},clip,width=\textwidth, page=2]{figures/ABC_26628.pdf}
     \end{subfigure}
     \hfill
     \begin{subfigure}[b]{0.12\textwidth}
        \centering
        \includegraphics[trim={0 0.750cm 0 1.5cm},clip,width=\textwidth, page=3]{figures/ABC_26628.pdf}
     \end{subfigure}
     \hfill
     \begin{subfigure}[b]{0.12\textwidth}
        \centering
        \includegraphics[trim={0 0.750cm 0 1.5cm},clip,width=\textwidth, page=4]{figures/ABC_26628.pdf}
     \end{subfigure}
     \hfill
     \begin{subfigure}[b]{0.12\textwidth}
        \centering
        \includegraphics[trim={0 0.750cm 0 1.5cm},clip,width=\textwidth, page=5]{figures/ABC_26628.pdf}
     \end{subfigure}
     \hfill
     \begin{subfigure}[b]{0.12\textwidth}
        \centering
        \includegraphics[trim={0 0.750cm 0 1.5cm},clip,width=\textwidth, page=6]{figures/ABC_26628.pdf}
     \end{subfigure}
     
     \begin{subfigure}[b]{0.12\textwidth}
        \centering
        \includegraphics[trim={0 0 0 0.75cm},clip, width=\textwidth, page=1]{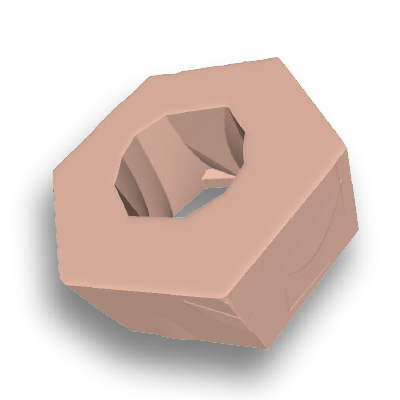}
     \end{subfigure}
     \hfill
     \begin{subfigure}[b]{0.12\textwidth}
        \centering
        \includegraphics[trim={0 0 0 0.75cm},clip,width=\textwidth, page=2]{figures/ABC_25446.pdf}
     \end{subfigure}
     \hfill
     \begin{subfigure}[b]{0.12\textwidth}
        \centering
        \includegraphics[trim={0 0 0 0.75cm},clip,width=\textwidth, page=3]{figures/ABC_25446.pdf}
     \end{subfigure}
     \hfill
     \begin{subfigure}[b]{0.12\textwidth}
        \centering
        \includegraphics[trim={0 0 0 0.75cm},clip,width=\textwidth, page=4]{figures/ABC_25446.pdf}
     \end{subfigure}
     \hfill
     \begin{subfigure}[b]{0.12\textwidth}
        \centering
        \includegraphics[trim={0 0 0 0.75cm},clip,width=\textwidth, page=5]{figures/ABC_25446.pdf}
     \end{subfigure}
     \hfill
     \begin{subfigure}[b]{0.12\textwidth}
        \centering
        \includegraphics[trim={0 0 0 0.75cm},clip,width=\textwidth, page=6]{figures/ABC_25446.pdf}
     \end{subfigure}
     
     \begin{subfigure}[b]{0.12\textwidth}
        \centering
        \includegraphics[trim={0 0 0 1.7cm},clip,width=\textwidth, page=1]{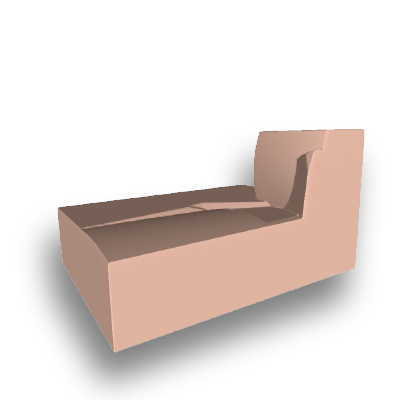}
     \end{subfigure}
     \hfill
     \begin{subfigure}[b]{0.12\textwidth}
        \centering
        \includegraphics[trim={0 0 0 1.7cm},clip,width=\textwidth, page=2]{figures/ABC_769.pdf}
     \end{subfigure}
     \hfill
     \begin{subfigure}[b]{0.12\textwidth}
        \centering
        \includegraphics[trim={0 0 0 1.7cm},clip,width=\textwidth, page=3]{figures/ABC_769.pdf}
     \end{subfigure}
     \hfill
     \begin{subfigure}[b]{0.12\textwidth}
        \centering
        \includegraphics[trim={0 0 0 1.7cm},clip,width=\textwidth, page=4]{figures/ABC_769.pdf}
     \end{subfigure}
     \hfill
     \begin{subfigure}[b]{0.12\textwidth}
        \centering
        \includegraphics[trim={0 0 0 1.7cm},clip,width=\textwidth, page=5]{figures/ABC_769.pdf}
     \end{subfigure}
     \hfill
     \begin{subfigure}[b]{0.12\textwidth}
        \centering
        \includegraphics[trim={0 0 0 1.7cm},clip,width=\textwidth]{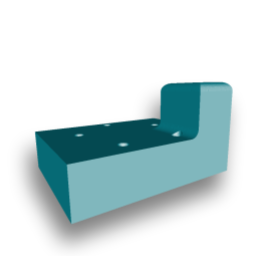}
     \end{subfigure}
     
     \begin{subfigure}[b]{0.12\textwidth}
        \centering
        \includegraphics[trim={0 0 0 1cm},clip,width=\textwidth, page=1]{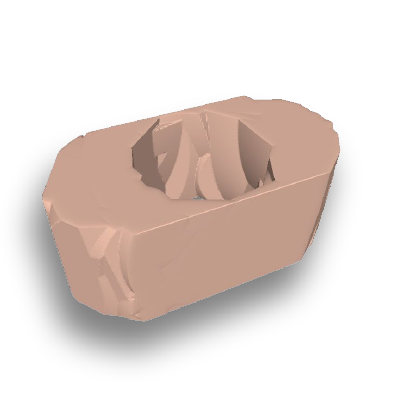}
        \caption*{CSG-Stump~[\textcolor{cvprblue}{29}]}
     \end{subfigure}
     \hfill
     \begin{subfigure}[b]{0.12\textwidth}
        \centering
        \includegraphics[trim={0 0 0 1cm},clip,width=\textwidth, page=2]{figures/ABC_36622.pdf}
        \caption*{ExtrudeNet~[\textcolor{cvprblue}{30}]}
     \end{subfigure}
     \hfill
     \begin{subfigure}[b]{0.12\textwidth}
        \centering
        \includegraphics[trim={0 0 0 1cm},clip,width=\textwidth, page=3]{figures/ABC_36622.pdf}
        \caption*{CAPRI-Net~[\textcolor{cvprblue}{37}]}
     \end{subfigure}
    \hfill
     \begin{subfigure}[b]{0.12\textwidth}
        \centering
        \includegraphics[trim={0 0 0 1cm},clip,width=\textwidth, page=4]{figures/ABC_36622.pdf}
        \caption*{SECAD-Net~[\textcolor{cvprblue}{23}]}
     \end{subfigure}
     \hfill
     \begin{subfigure}[b]{0.12\textwidth}
        \centering
        \includegraphics[trim={0 0 0 1cm},clip,width=\textwidth, page=5]{figures/ABC_36622.pdf}
        \caption*{CNC-Net~(Ours)}
     \end{subfigure}
     \hfill
     \begin{subfigure}[b]{0.12\textwidth}
        \centering
        \includegraphics[trim={0 0 0 1cm},clip,width=\textwidth, page=6]{figures/ABC_36622.pdf}
        \caption*{Target~(GT)}
     \end{subfigure}
        \caption{
        \textbf{Qualitative results on ABC~[\textcolor{cvprblue}{16}] dataset.} 
         All shapes are visualized using marching cubes~(MC) with $256$ resolution.}
        \label{fig:qual_abc_supp}
        \vspace{-4mm}
\end{figure*}     

\begin{figure*}[h]
     \centering
     \begin{subfigure}[b]{0.12\linewidth}
        \centering
        \includegraphics[trim={0 0cm 0 0},clip,width=\linewidth, page=1]{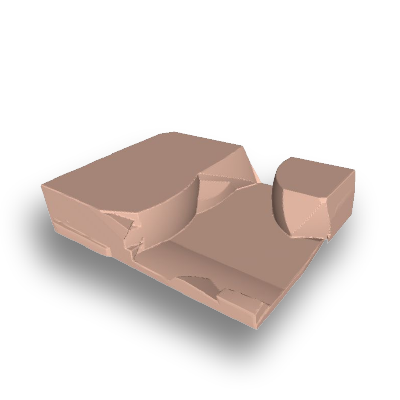}
     \end{subfigure}
     \hfill
     \begin{subfigure}[b]{0.12\linewidth}
        \centering
        \includegraphics[trim={0 0cm 0 0},clip,width=\linewidth, page=2]{figures/ShapeNet_3.pdf}
     \end{subfigure}
     \hfill
     \begin{subfigure}[b]{0.12\linewidth}
        \centering
        \includegraphics[trim={0 0cm 0 0},clip,width=\linewidth, page=3]{figures/ShapeNet_3.pdf}
     \end{subfigure}
     \hfill
     \begin{subfigure}[b]{0.12\linewidth}
        \centering
        \includegraphics[trim={0 0cm 0 0},clip,width=\linewidth, page=4]{figures/ShapeNet_3.pdf}
     \end{subfigure}
     \hfill
     \begin{subfigure}[b]{0.12\linewidth}
        \centering
        \includegraphics[trim={0 0cm 0 0},clip,width=\linewidth, page=5]{figures/ShapeNet_3.pdf}
     \end{subfigure}
     \hfill
     \begin{subfigure}[b]{0.12\linewidth}
        \centering
        \includegraphics[trim={0 0cm 0 0},clip,width=\linewidth]{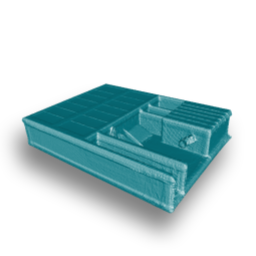}
     \end{subfigure}
     
     \begin{subfigure}[b]{0.12\linewidth}
        \centering
        \includegraphics[trim={0 0 0 1cm},clip,width=\linewidth, page=1]{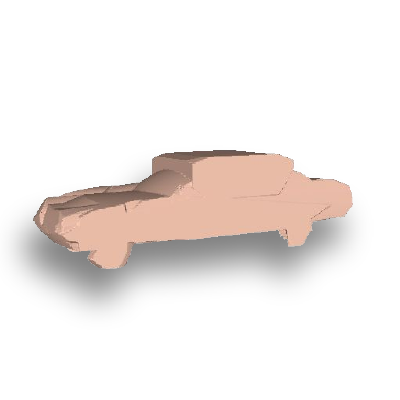}
     \end{subfigure}
     \hfill
     \begin{subfigure}[b]{0.12\linewidth}
        \centering
        \includegraphics[trim={0 0 0 1cm},clip,width=\linewidth, page=2]{figures/ShapeNet_4.pdf}
     \end{subfigure}
     \hfill
     \begin{subfigure}[b]{0.12\linewidth}
        \centering
        \includegraphics[trim={0 0 0 1cm},clip,width=\linewidth, page=3]{figures/ShapeNet_4.pdf}
     \end{subfigure}
     \hfill
     \begin{subfigure}[b]{0.12\linewidth}
        \centering
        \includegraphics[trim={0 0 0 1cm},clip,width=\linewidth, page=4]{figures/ShapeNet_4.pdf}
     \end{subfigure}
     \hfill
     \begin{subfigure}[b]{0.12\linewidth}
        \centering
        \includegraphics[trim={0 0 0 1cm},clip,width=\linewidth, page=5]{figures/ShapeNet_4.pdf}
     \end{subfigure}
     \hfill
     \begin{subfigure}[b]{0.12\linewidth}
        \centering
        \includegraphics[trim={0 0 0 1cm},clip,width=\linewidth]{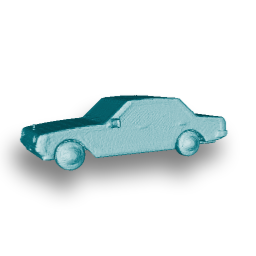}
     \end{subfigure}

     \begin{subfigure}[b]{0.12\linewidth}
        \centering
        \includegraphics[trim={0 0 0 2cm},clip,width=\linewidth, page=1]{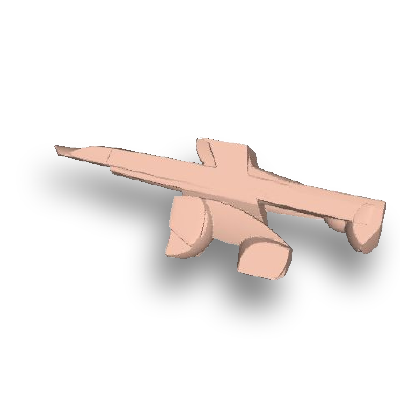}
     \end{subfigure}
     \hfill
     \begin{subfigure}[b]{0.12\linewidth}
        \centering
        \includegraphics[trim={0 0 0 2cm},clip,width=\linewidth, page=2]{figures/ShapeNet_9.pdf}
     \end{subfigure}
     \hfill
     \begin{subfigure}[b]{0.12\linewidth}
        \centering
        \includegraphics[trim={0 0 0 2cm},clip,width=\linewidth, page=3]{figures/ShapeNet_9.pdf}
     \end{subfigure}
     \hfill
     \begin{subfigure}[b]{0.12\linewidth}
        \centering
        \includegraphics[trim={0 0 0 2cm},clip,width=\linewidth, page=4]{figures/ShapeNet_9.pdf}
     \end{subfigure}
     \hfill
     \begin{subfigure}[b]{0.12\linewidth}
        \centering
        \includegraphics[trim={0 0 0 2cm},clip,width=\linewidth, page=5]{figures/ShapeNet_9.pdf}
     \end{subfigure}
     \hfill
     \begin{subfigure}[b]{0.12\linewidth}
        \centering
        \includegraphics[trim={0 0 0 2cm},clip,width=\linewidth]{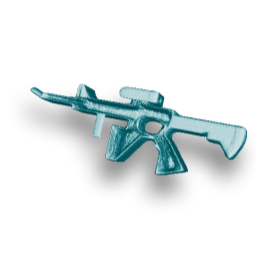}
     \end{subfigure}
     
     \begin{subfigure}[b]{0.12\linewidth}
        \centering
        \includegraphics[trim={0 0 0 1cm},clip,width=\linewidth, page=1]{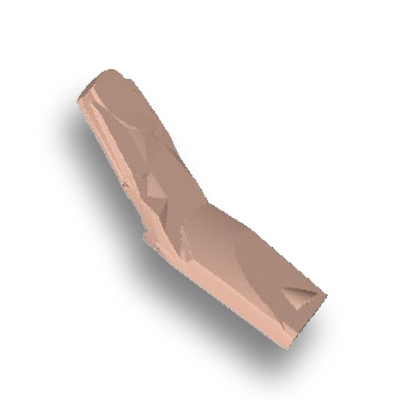}
        \caption*{ CSG-Stump~[\textcolor{cvprblue}{29}]}
     \end{subfigure}
     \hfill
    \begin{subfigure}[b]{0.12\linewidth}
        \centering
        \includegraphics[trim={0 0 0 1cm},clip,width=\linewidth, page=1]{figures/ShapeNet_12.pdf}
        \caption*{ExtrudeNet~[\textcolor{cvprblue}{30}]}
     \end{subfigure}
     \hfill
     \begin{subfigure}[b]{0.12\linewidth}
        \centering
        \includegraphics[trim={0 0 0 1cm},clip,width=\linewidth, page=3]{figures/ShapeNet_12.pdf}
        \caption*{CAPRI-Net~[\textcolor{cvprblue}{37}]}
     \end{subfigure}
    \hfill
     \begin{subfigure}[b]{0.12\linewidth}
        \centering
        \includegraphics[trim={0 0 0 1cm},clip,width=\linewidth, page=4]{figures/ShapeNet_12.pdf}
        \caption*{SECAD-Net~[\textcolor{cvprblue}{23}]}
     \end{subfigure}
     \hfill
     \begin{subfigure}[b]{0.12\linewidth}
        \centering
        \includegraphics[trim={0 0 0 1cm},clip,width=\linewidth, page=5]{figures/ShapeNet_12.pdf}
        \caption*{CNC-Net~(Ours)}
     \end{subfigure}
     \hfill
     \begin{subfigure}[b]{0.12\linewidth}
        \centering
        \includegraphics[trim={0 0 0 1cm},clip,width=\linewidth]{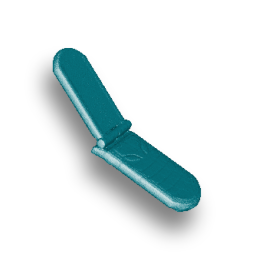}
        \caption*{Target~(GT)}
     \end{subfigure}
        \caption{
        \textbf{Qualitative results on ShapeNet~[\textcolor{cvprblue}{3}] dataset.} 
        All shapes are visualized using marching cubes~(MC) with $256$ resolution.}
        \label{fig:qual_shapenet_supp}
        \vspace{-4mm}
\end{figure*}

\section{Milling paths}
In Sec.~4.4.1 of the main manuscript,  we present visualizations of the milling paths for some randomly selected samples from the ABC~[\textcolor{cvprblue}{16}] dataset.
In \cref{fig:path_supp}, we further provide the generated paths of the first four subsequent operations for more samples from the ABC~[\textcolor{cvprblue}{16}] dataset.
The results demonstrate the capacity of our method to intelligently learn paths in a self-supervised manner, eliminating the necessity for ground truth paths.

\section{Zero-shot Vs. fine-tuning}
In Sec.~4.4.2 of the main manuscript, we conduct a quantitative comparison between results achieved through fine-tuning and those obtained via zero-shot learning.
To further analyze the differences between the results obtained by these two training strategies, we present a visual comparison of our outcomes on the ABC~[\textcolor{cvprblue}{16}] dataset, as illustrated in \cref{fig:zero_vs_fine_supp}.
The qualitative results indicate that our zero-shot learning method can effectively reproduce the general shapes of target objects. 
However, leveraging a pre-trained model and fine-tuning it for each sample notably enhances the reproduction of finer details. 
The zero-shot capability significantly enhances the practicality of our method, particularly when tackling target objects that differ substantially from shapes in the training dataset.

\vspace{-1mm}
\section{Effect of operations}
In Sec.~4.4.4 of the main manuscript, we quantitatively demonstrate the impact of each designed operation on achieving enhanced performance.
Furthermore, we visually represent their effects using some samples from the ABC~[\textcolor{cvprblue}{16}] dataset as shown in \cref{fig:wo}.
Our qualitative results also underscore the necessity of the milling and rotation operations in forming the general shape of the target object.
On the contrary, the drilling operation significantly improves precision in production.
Specifically, in the case of Wo $\mathcal{O^M}$ as depicted in \cref{fig:wo}, the method attempts to generate the shape by creating a hole, resulting in a shape that remains notably distant from the intended target shape.
On the other hand, the method Wo $\mathcal{R}ot$ either removes some parts of the target object~(first row) or leaves all parts intact~(second row).
Furthermore, while the method Wo $\mathcal{O^D}$ reproduces shapes resembling the targe shapes,  it still fails to capture finer details.

\vspace{-1mm}
\section{More qualitative results}
We expand our comparative analysis by offering additional visual comparisons between our CNC-Net method and SOTA 3D CAD reconstruction approaches, including CSG-Stump~[\textcolor{cvprblue}{29}], ExtrudeNet~[\textcolor{cvprblue}{30}], CAPRI-Net~[\textcolor{cvprblue}{37}], and SECAD-Net~[\textcolor{cvprblue}{23}], assessed across both the ABC~[\textcolor{cvprblue}{16}] and the ShapeNet~[\textcolor{cvprblue}{3}] datasets.
We illustrate these comparisons for the ABC and the ShapeNet datasets in \cref{fig:qual_abc_supp} and \cref{fig:qual_shapenet_supp}, respectively.